\documentclass[fleqn,10pt]{wlscirep}
\usepackage[utf8]{inputenc}
\usepackage[T1]{fontenc}
\usepackage{booktabs}
\usepackage{hyperref}
\usepackage{float}
\usepackage{xcolor}
\usepackage{caption}
\usepackage{subcaption}
\usepackage{multirow}
\usepackage{makecell}

\title{

A Scalable Workflow to Build Machine Learning Classifiers with Clinician-in-the-Loop to Identify Patients in Specific Diseases}

\author[1,3,=]{Jingqing Zhang}
\author[3,=]{Atri Sharma}
\author[3,=]{Luis Bolanos}
\author[3]{Tong Li}
\author[3]{Ashwani Tanwar}
\author[3]{Vibhor Gupta}
\author[1,2,3]{Yike Guo}
\affil[1]{Data Science Institute, Imperial College London, London, SW7 2AZ, UK}
\affil[2]{Hong Kong Baptist University, Hong Kong SAR, China}
\affil[3]{Pangaea Data Limited, UK, USA}

\affil[*]{Corresponding to y.guo@imperial.ac.uk}

\affil[=]{These authors contributed equally to this work}

\begin{abstract}

\textbf{Background:}
Clinicians and researchers may rely on medical coding systems such as International Classification of Diseases (ICD) to identify patients with specific diseases from Electronic Health Records (EHRs). However, due to the lack of detail and specificity as well as a probability of miscoding, recent studies suggest the ICD codes often cannot characterise patients accurately for specific diseases in real clinical practice, and as a result, using them to find patients for studies or trials can result in high failure rates and missing out on uncoded patients. Manual inspection of all patients at scale is not feasible as it is highly costly and slow.

\textbf{Methodology:}
This paper proposes a scalable workflow which leverages both structured data and unstructured textual notes from EHRs with techniques including Natural Language Processing (for phenotyping), AutoML and Clinician-in-the-Loop mechanism to build machine learning classifiers to identify patients at scale with given diseases, especially those who might currently be miscoded or missed by ICD codes. 

\textbf{Results:}
Case studies in the MIMIC-III dataset were conducted where the proposed workflow demonstrates a higher classification performance in terms of F1 scores compared to simply using ICD codes on gold testing subset to identify patients with Ovarian Cancer (0.901 vs 0.814), Lung Cancer (0.859 vs 0.828), Cancer Cachexia (0.862 vs 0.650), and Lupus Nephritis (0.959 vs 0.855). Also, the proposed workflow that leverages unstructured notes consistently outperforms the baseline that uses structured data only with an increase of F1 (Ovarian Cancer 0.901 vs 0.719, Lung Cancer 0.859 vs 0.787, Cancer Cachexia 0.862 vs 0.838 and Lupus Nephritis 0.959 vs 0.785). Experiments on the large testing set also demonstrate the proposed workflow can find more patients who are miscoded or missed by ICD codes. Moreover, interpretability studies are also conducted to clinically validate the top impact features behind the decision-making of the classifiers.

\textbf{Conclusions:}
The proposed workflow can more accurately identify patients with specific diseases than simply using ICD codes. We also find the phenotypic features extracted from unstructured textual notes are beneficial for better accuracy and interpretability of classifiers. Moreover, the proposed workflow is scalable to other diseases and use cases as Clinician-in-the-Loop and AutoML enable rapid configuration of new machine learning classifiers. 
\end{abstract}

\newpage

\begin{document}
\newpage
\flushbottom
\maketitle
%
%
\thispagestyle{empty}



\section{Introduction}
\label{sec:introduction}

The accumulation of healthcare data has recently reached unprecedented levels, with the National Health Service (NHS) in the UK recording approximately 560 million patient interactions each year. \cite{NHSActivity} The collection of this information is critical for research and development in the medical domain, such as developing new drugs, treatments or therapies; understanding the characteristics and comorbidities of rare diseases; techniques for the earlier detection of certain diseases. \cite{Dash2019, Vamathevan2019, Park2021, Hassaine2020} For all these use cases, it is critical to identify the target patients first at scale, either for recruitment into studies, or a deeper analysis of the data of patients and diseases. 

A commonly used method to identify such target patients is to use medical coding systems like International Classification of Diseases (ICD). However, several studies have highlighted flaws in their real-world application which make them unsuitable for classifying patients, especially with rare and/or highly specific diseases in real practice. One study \cite{Wockenfuss2009} conducted a thorough investigation involving 200 General Practitioners (GPs) in the Saxony region to evaluate the inter-rater agreement between two coders on the same patient’s health record, and came to the conclusion that ICD was not reliable in primary care. Another large-scale survey \cite{southern_etal} of 250 clinicians from 12 countries reported that the majority of the respondents found that ICD codes had no clustering mechanisms, lacked specificity, had no terms for describing complications or adverse events, and had vague term definitions. A more recent study \cite{Snyder2017} evaluated the accuracy of ICD-9-CM codes in determining the genotypes for Sickle Cell Disease for healthcare quality studies, and found that ICD codes displayed an accuracy as low as 23\% for certain SCD genotypes, making them unsuitable for research purposes. A similar study for Crohn's disease and diabetes \cite{Horsky2018} analysed the accuracy and completeness of ICD codes in ambulatory visits by analysing the coding performed by 23 clinicians and found that over 25\% of the appropriate codes were not recorded and omitted, leading to an incorrect and incomplete characterisation of the patients’ conditions. ICD codes are also prone to miscoding as a recent study conducted in the context of pulmonary embolisms \cite{Burles2017} found that up to 18\% of patients coded with ICD-10 were false positives, and 9\% were false negatives. The authors attributed these errors to the vague documentation for physicians responsible for assigning codes. Overall, as the ICD codes may not be reliable for a number of diseases and manual inspection of thousands of or even millions of patients by clinical experts is highly costly and time-consuming, an automatic pipeline with machine learning and specific expert guidelines can be useful to standardise, scale-up and accelerate the identification of patients with rare and complex diseases. 

Recent studies have demonstrated the feasibility of using machine learning (including deep learning) models to classify patients with specific diseases \cite{xu2020identifying,liu2021evaluating}. The majority of such previous studies utilise structured data (e.g., blood test results, laboratory data, demographics) without using unstructured textual data (e.g., clinical notes, discharge summaries, radiology reports) \cite{wu2019prediction,sekelj2021detecting}. For example, structured variables collected from renal biopsies are used to predict the onset of renal flares in known cases of Lupus Nephritis \cite{Chen2021} and gene expression data to predict lupus disease \cite{Kegerreis2019}. However, the unstructured textual data contains rich clinical information and studies show that up to 80\% of data in EHRs is unstructured \cite{Wei2015, Kong2019}. Moreover, despite the flaws of ICD code assignment in real clinical practice, recent studies still use ICD codes of Electronic Health Records (EHRs) as target labels for the training and evaluation of machine learning models to predict disease diagnosis \cite{xu2019multimodal,zhang-etal-2020-bert} but we believe the involvement of clinicians in a study to revise patient diagnosis labels is necessary. 

Therefore, we propose a machine learning workflow with the Clinician-in-the-Loop mechanism that leverages both structured data and unstructured textual data from EHRs to find (1) critical clinical features that help characterise a disease, especially for rare diseases which are not well defined and understood; (2) relevant patients meeting the precise criteria of the disease. More precisely, to leverage the rich clinical information from unstructured textual data, we incorporate a state-of-the-art phenotyping algorithm with Natural Language Processing (NLP) to extract phenotypic features \footnote{In this study, the word ``phenotype'' refers to deviations from normal morphology, physiology, or behaviour, as defined by the study \cite{robinson2012deep}.} of patients. The phenotypic features are then enriched with the features from structured data for predictive modelling by Automated Machine Learning (AutoML) to classify the patients. AutoML is a framework for efficient and automated feature, classifier type, and hyper-parameter selection of machine learning classifiers. The classifiers are iteratively improved with gold diagnosis labels and feedback on features from clinicians via the Clinician-in-the-Loop mechanism. 
We evaluate the proposed workflow on four diseases including Ovarian Cancer, Lung Cancer, Cancer Cachexia and Lupus Nephritis and show superior performance to baseline methods. 

Different from the work \cite{yu2018enabling}, the proposed workflow does not use ICD codes as input features given their flaws. In addition, we build the disease classifiers with AutoML which use phenotypes as input features because the diagnosis of diseases may rely on a combination of multiple phenotypes (such as the American College of Rheumatology Classification Criteria \cite{SLE_Criteria} for systemic lupus erythematosus (SLE)) even though the diseases themselves may not be explicitly mentioned in clinical notes. This also suggests that using phenotyping NLP algorithms like Clinphen \cite{deisseroth2019clinphen}, NCBO \cite{jonquet2009ncbo}, cTAKES \cite{Savova2010}, MetaMap and \cite{Aronson2010} solely without classifiers is not sufficient to perform inference and identify patients with specific diseases.

In summary, our \textbf{main contributions} of this work are as follows: 

\begin{enumerate}
    \item We propose a scalable workflow with phenotyping, AutoML and Clinician-in-the-Loop to identify patients of specific diseases accurately which can be potentially applied to multiple diseases. 
    
    \item We demonstrate that using phenotypic features from unstructured textual data leads to higher accuracy of patient identification and better interpretability of the profile of the patients at risk.
    
    \item We demonstrate that AutoML is an efficient and effective solution to build machine learning classifiers for patient identification.
    
    \item We also demonstrate that the Clinician-in-the-Loop mechanism improves the training set and incorporates clinically relevant features into classifiers which leads to better accuracy and interpretability of classifiers.
    
    
\end{enumerate}

\begin{figure}[!t]
\centering
\centering
\includegraphics[width=0.8\linewidth]{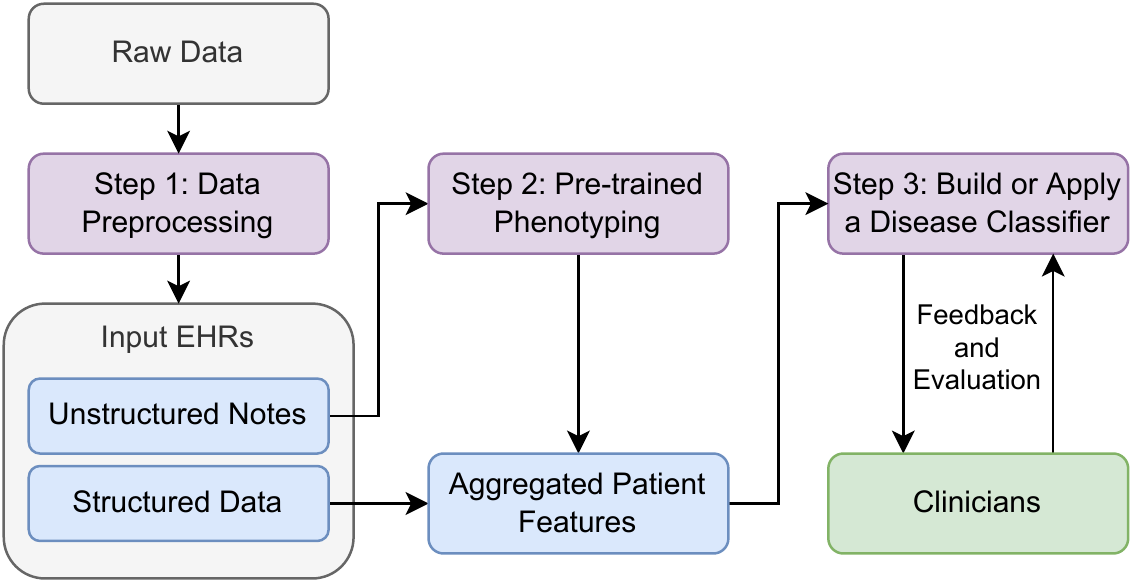}
\caption{Given one specific disease, this figure illustrates the proposed workflow to build, apply and evaluate a disease classifier with clinician-in-the-loop to identify patients at risk of the specific disease.}
\label{fig:workflow}
\end{figure}

\section{Methodology}
\label{sec:methodology}

In this section, we will introduce the proposed workflow which is shown in Figure \ref{fig:workflow}. As an overview, the workflow mainly consists of three steps modules. First, the data preprocessing module gathers raw data from the database for each disease and prepare unstructured textual data (such as discharge summaries, nursing notes, and radiology reports) and structured data (such as blood test results, temperature, and heart rate) from the raw data. Second, we run a pre-trained state-of-the-art phenotyping algorithm to extract phenotypic features of patients from unstructured textual data \cite{zhang-etal-2021-self}. Third, the phenotypic features are combined with structured data together as the aggregated features of patients which will be fed into the disease classifier. The disease classifier works as a binary classifier for each specific disease which is built based on gold diagnosis labels and feedback from clinicians on the training set and then evaluated on the testing set. The three steps will be elaborated in the following subsections.

\subsection{Data Preprocessing}
\label{subsec:data_prepro}

In this study, we use the publicly available EHR dataset MIMIC-III \cite{johnson2016mimic} for better reproducibility. For the unstructured textual data, we collect discharge summaries which are provided to patients at the end of admissions and contain extensive clinical information, especially phenotypes about the patients. For the structured data, we follow the practice \cite{harutyunyan2019multitask} and collect 17 clinical features such as weight, temperature, heart rate, blood pressure and respiratory rate. The structured data is aggregated across the entire admission of a patient and mean imputation is applied to missing values. The units for the same structured feature are unified, for example, Fahrenheit to Celsius for the feature temperature.

\subsection{Pre-trained Context-aware Phenotyping NLP Algorithm}
\label{sec:phenotyping}

The unstructured clinical notes, such as discharge summaries, nursing notes and radiology reports, are rich in phenotype information as the clinicians naturally describe phenotypic abnormalities of patients in the narratives of notes. Previous studies have demonstrated leveraging the phenotype information to improve the understanding of disease diagnosis, disease pathogenesis, patient outcomes and genomic diagnostics \cite{aerts2006gene,son2018deep,Liu2019,deisseroth2019clinphen,Xu2020,zhang2021clinical}, and subsequently, the automatic phenotype annotation from clinical notes has become an important task in clinical Natural Language Processing (NLP). 

In the proposed workflow, we use the state-of-the-art phenotyping NLP algorithm developed by the recent work \cite{zhang-etal-2021-self} to extract a list of phenotypes from unstructured clinical notes. The algorithm uses Human Phenotype Ontology (HPO) \cite{DBLP:journals/nar/0001GMCLVDBBBCC21} to standardize over 15,000 phenotype concepts and leverages self-supervised pre-training techniques, contextualized word embeddings by the Transformer model \cite{vasvani:nips:2018} and data augmentation techniques (paraphrasing and synthetic text generation) to capture names (like hypertension), synonyms (like high blood pressure), abbreviations (like HTN) and, more importantly, contextual synonyms of phenotypes. For example, descriptive phrases like ``rise in blood pressure'', ``blood pressure is increasing'' and ``BP of 140/90'' are considered as contextual synonyms of \textit{Hypertension (HP:0000822)} and finding such contextual synonyms require an understanding of context semantics. As a result of the contextual detection of phenotype, the phenotyping NLP algorithm \cite{zhang-etal-2021-self} demonstrates superior performance to baseline phenotyping algorithms including Clinphen \cite{deisseroth2019clinphen}, NCBO \cite{jonquet2009ncbo}, cTAKES \cite{Savova2010}, MetaMap \cite{Aronson2010}, MetaMapLite \cite{demner2017metamaplite}), NCR \cite{arbabi2019ncr}, MedCAT \cite{kraljevic2019medcat} and fine-tuned BERT-based models \cite{devlin-etal-2019-bert,biobert,alsentzer-etal-2019-publicly,beltagy-etal-2019-scibert}. 

\subsection{Disease Classifier}

In the proposed workflow as shown in Figure \ref{fig:workflow}, the list of phenotypes that are extracted from unstructured clinical notes by the phenotyping NLP algorithm will be combined with the 17 structured clinical features (as mentioned in section \ref{subsec:data_prepro}) together as the aggregated features of patients. For instance, the aggregated features of a patient may have a list of phenotypes like weight loss (HP:0001824) and malnutrition (HP:0004395) and structured clinical features like temperature 36.6 and heart rate 86. The aggregated clinical features of patients will be then used as input features in the subsequent step which leverages Automated Machine Learning (AutoML) and Clinician-in-the-Loop mechanism to build a disease classifier on the training set and evaluate the disease classifier on the testing set.

\subsubsection{Automated Machine Learning (AutoML)}
\label{subsec:automl}

To build a disease classifier, we use Automated Machine Learning (AutoML) \cite{he2021automl} which provides efficient and scalable solutions to build machine learning classifiers with minimal human assistance. In convention, building a machine learning model (including deep learning) involves a pipeline of steps that require human expertise including feature engineering, model selection, model architecture search, hyper-parameter tuning, model evaluation and more. Given the complexity of the steps, it is usually time-consuming to manually find the best configuration of a machine learning classifier and requires expertise from machine learning engineers. Therefore, AutoML is designed to automate the pipeline within a time budget by using strategies such as grid search, Bayesian Optimization and meta-learning.

More specifically, we develop an AutoML framework based on the HpBandSter library \cite{falkner-icml-18}. This framework automatically searches for the most suitable classifier model from candidates including Support Vector Machine (SVM), Random Forest, Gradient Boosting, Logistic Regression, and Multi-layer Perceptron as well as their respective hyper-parameters. The AutoML framework also incorporates feature engineering from scikit-learn \cite{scikit-learn} which, in practice, automatically selects top features from the aggregated features of patients. For example, weight loss (HP:0001824) is selected as the top indicative feature for Cancer patients with Cachexia as shown in Figure \ref{fig:cachexia_ind_shap}.

\subsubsection{Clinician-in-the-Loop Mechanism}
\label{subsec:clinicianloop}

Building disease classifiers by machine learning algorithms is restricted by two challenges: (1) lack of accurate and gold standard labels from medical experts to train and evaluate the models, for example, ICD codes in EHR database may not be accurate as discussed in Section \ref{sec:introduction} and manually labelling EHRs at scale is highly costly; (2) lack of interpretability and clinical validity behind the decision made by the models. Therefore, we leverage Clinician-in-the-Loop mechanism, by which clinicians can iteratively work with the models and provide feedback for enhancement, to address the two challenges by collecting gold diagnosis labels for patients and validating clinically relevant features of diseases with clinicians.

In the training stage, we first request three clinicians to create gold diagnosis labels for patients with consensus. The annotation guideline instructs the clinicians to provide a binary label for each patient on whether the patient suffers from, has a recent history of, or is at risk of a specific disease or not. The labels provided by the clinicians are collected as the gold diagnosis labels, which are used as the targets to train disease classifiers by AutoML.

Meanwhile, we also iteratively request the clinicians to validate the top features that a disease classifier relies on to identify patients. The importance of features is quantified by SHAP values \cite{SHAP} which measure how much a feature positively or negatively impacts the decision of a classifier. The features that have the highest absolute SHAP values are taken as top features and sent to the clinicians to assess their clinical relevance to the disease. The clinically irrelevant features will be removed from the input features of patients while the relevant features will be retained. The removal of such statistically correlated but clinically irrelevant features can potentially help prevent overfitting and improve the interpretability of model predictions. The final disease classifier is trained after several iterations of feature validation until the model performance is stable (typically within 3 iterations in practice).

In the testing stage, after the final disease classifier is obtained, it will be applied to the testing data to predict the patients at high risk of the disease. The evaluation method will be explained in Section \ref{sec:evaluation_method}.

\subsection{Dataset and Gold Labelling}

We use 58,976 Electronic Health Records (EHRs) from MIMIC-III database \cite{johnson2016mimic} which corresponds to 58,976 admissions from 46,520 unique patients in Intensive Care Units (ICU). Each EHR in MIMIC-III is assigned with relevant ICD-9 codes. The EHRs are randomly split into the training set with 29,637 EHRs and the testing set with 29,339 EHRs, and the workflow makes one prediction for each EHR. Please note the EHRs of the same unique patients will be all in either the training set or the testing set to avoid data leakage.

    
    


We conduct experiments on four diseases including Ovarian Cancer, Lung Cancer, Cancer Cachexia, and Lupus Nephritis. The definitions of these diseases based on ICD-9 codes in this study are shown in Table \ref{tab:training_dataset_ICD}. The ICD-based criteria are then used as ``noisy'' labels for EHRs and as a baseline method to identify patients. 

\begin{table}[H]
\centering
\begin{tabular}{@{}c|c|c@{}}
\toprule
\textbf{Disease Name}   & \textbf{ICD-9 Codes for Positive Cohort}      & \textbf{ICD-9 Codes for Negative Cohort} \\ \midrule
Ovarian Cancer          & 183.0                                         & otherwise                          \\
Lung Cancer             & any subcode of 162                            & otherwise                          \\ 
Cancer Cachexia         & any subcode 140–239 and (799.3 or 799.4) & any subcode 140–239 but not 799.3 nor 799.4 \\ 
Lupus Nephritis         & 710.0 and any subcode of 580                  & otherwise                          \\ \bottomrule
\end{tabular}
\caption{The ICD code-based criteria to define positive and negative cohorts for each disease in the training and testing set. Please note the Cancer Cachexia setup requires patients to have cancer as the background condition for both positive and negative cohorts.}
\label{tab:training_dataset_ICD}
\end{table}

\begin{table}[H]
\centering
\begin{tabular}{@{}c|cc|cc|cc|cc@{}}
\toprule
& \multicolumn{2}{c|}{\textbf{\makecell{Entire Training Set \\ by ICD Labels}}} & \multicolumn{2}{c|}{\textbf{\makecell{Training Subset \\ by Gold Labels}}} & \multicolumn{2}{c|}{\textbf{\makecell{Entire Testing Set \\ by ICD Labels}}}& \multicolumn{2}{c}{\textbf{\makecell{Testing Subset \\ by Gold Labels}}} \\
\textbf{Disease Name} & Positive & Negative & Positive & Negative & Positive & Negative & Positive & Negative \\ \midrule
Ovarian Cancer  & 43  & 29,594 & 57 & 43 & 38  & 29,301 & 52 & 48    \\
Lung Cancer     & 586 & 29,051 & 62 & 38 & 585 & 28,754 & 66 & 34    \\
Cancer Cachexia* & 42 & 4,449 & 59 & 33 & 51 & 4,370 & 70 & 27   \\ 
Lupus Nephritis & 86  & 29,551 & 42 & 58 & 63  & 29,276 & 62 & 38    \\ \bottomrule
\end{tabular}
\caption{The number of positive and negative EHRs in the entire training set and entire testing set based on the ICD-based criteria defined by Table \ref{tab:training_dataset_ICD}, as well as the training subset and testing subset which have gold diagnosis labels created via the Clinician-in-the-Loop mechanism. Please note the training (testing) subset with gold labels is fully included by the entire training (testing) set. The disease classifiers will be trained to differentiate the positive and negative EHRs. * The number of EHRs with gold labels in the training and testing subset for Cancer Cachexia is fewer than 100 because a few patients who do not have cancer as the background condition are discarded.}
\label{tab:gold_cohort}
\end{table}

Table \ref{tab:gold_cohort} summarises the number of positive and negative EHRs in the entire training and testing set based on the ICD-based criteria which also shows that the prevalence of these diseases is low in MIMIC-III. However, the ICD codes cannot be used as the gold diagnosis labels given their multiple flaws, so we use the Clinician-in-the-Loop mechanism to create the gold diagnosis labels for EHRs. As manually creating large-scale gold labels on entire MIMIC-III datasets with clinicians is not feasible in practice, only small-scale EHRs can be selected for gold labelling and ideally, the selected EHRs should be relatively balanced (i.e. around half EHRs are positive for the disease and the other half are negative).

To create the relatively balanced gold set, an initial yet imperfect disease classifier (a Random Forest classifier \cite{Breiman2001}) is first trained solely based on the entire training set by using ICD codes as the learning target (i.e., positive and negative cohorts are decided by ICD criteria in Table \ref{tab:training_dataset_ICD}). Then, 100 EHRs are selected from the training set by randomly sampling 25 EHRs from each of the following four patient groups: (1) EHRs that are predicted/labelled as positive by the initial classifier and ICD; (2) EHRs that are predicted as positive by the initial classifier but labelled as negative by ICD; (3) EHRs that are predicted as negative by the initial classifier but labelled as positive by ICD; (4) EHRs that are predicted/labelled as negative by the initial classifier and ICD. Similarly, another 100 EHRs are selected from the entire testing set by applying the same initial classifier. As opposed to selecting EHRs based on ICD codes directly, using the initial classifier with the four patient groups encourages finding EHRs that are mislabelled by ICD codes. Please note the initial classifier is preliminary and only used to select these 200 EHRs. Next, the 200 EHRs are manually labelled via the Clinician-in-the-Loop mechanism following the annotation guideline. Table \ref{tab:gold_cohort} also summarises the statistics of the training subset and testing subset which have gold diagnosis labels.

\subsection{Evaluation Methods and Baselines}
\label{sec:evaluation_method}

We consider two evaluation methods for the disease classifier. The first evaluation method is to compare the disease classifier created by the proposed workflow with baseline methods on the gold testing set. We consider the following baseline methods. (1) ICD codes: we use the inclusion and exclusion criteria based on ICD codes as defined by Table \ref{tab:training_dataset_ICD} to classify EHRs. (2) Structured data only: using the 17 clinical features from structured data only as patient features to AutoML without considering phenotypic features in unstructured notes. (3) Structured + NCR: To compare with other phenotyping NLP algorithms, we also aggregate the structured clinical features with the phenotypes that are extracted from clinical notes by NCR \cite{arbabi2019ncr} and then we use AutoML to build disease classifiers. (4) Structured + ClinicalBERT: this is similar to the previous baseline method but the phenotypes are extracted by ClinicalBERT which are fine-tuned for phenotyping following the work \cite{zhang-etal-2021-self}. On the gold testing set, we report Area Under the Curve of Receiver Operating Characteristic (AUC-ROC) and Area Under the Curve of Precision-Recall (AUC-PR) in addition to precision, recall, F1, and specificity which use 0.5 as the threshold. Please note AUC-ROC and AUC-PR are not applicable to ICD codes as the labels of ICD codes are discrete in binary (either 0 or 1) rather than continuous in probability.

As the gold testing set has only 100 EHRs in total for each disease, the insights from the first evaluation method may not be comprehensive. Therefore, we use the second evaluation method, which runs the disease classifier on the entire testing set which has 29,339 EHRs. We report the number of EHRs ($N_{\text{pred}}$ ) which are predicted as positive by the disease classifier and the estimated precision ($P_{\text{est}}$). To calculate the estimated precision, we first randomly select 100 EHRs that are predicted as positive and request validation from three clinicians with consensus if the 100 EHRs have true patients. The estimated precision is then computed by dividing the number of EHRs that have true patients by 100. Moreover, we estimate the number of actual positive EHRs that are found by the disease classifier by $N_{\text{pred}} \times P_{\text{est}}$ with rounding to the nearest integer. The second evaluation method is more challenging than the first evaluation method as the prevalence of the four diseases is very low in the entire testing set according to Table \ref{tab:gold_cohort}.

\section{Results}


\begin{table}[h]
\centering
\begin{tabular}{@{}c|c|ccccccc@{}}
\toprule
\textbf{Disease Name} & \textbf{Method} & \textbf{Precision} & \textbf{Recall} & \textbf{F1} & \textbf{Specificity} & \textbf{AUC-PR} & \textbf{AUC-ROC}  \\ \midrule 
\multirow{5}{*}{Ovarian Cancer} & ICD Codes   & 0.946 & 0.714 & 0.814 & 0.957 & 0.834 & 0.836  \\
& Structured Data Only   & 0.605 &	0.885 &	0.719 &	0.375 &	0.595 & 0.630 \\
& Structured + NCR & 0.797 & 0.981 & 0.879 & 0.729 & 0.792 & 0.855 \\
& Structured + ClinicalBERT & 0.788 & 1.000 & 0.881 & 0.708 & 0.788 & 0.854 \\
& Ours  & 0.847	& 0.962	& \textbf{0.901}	& 0.813	& 0.821	& \textbf{0.877} \\ \hline
\multirow{5}{*}{Lung Cancer} & ICD Codes   & 0.960 & 0.727 & 0.828 & 0.941 & 0.878 & \textbf{0.834} \\
& Structured Data Only   & 0.702 & 0.894 & 0.787 & 0.265 & 0.698 & 0.579 \\
& Structured + NCR & 0.780 & 0.970 & \textbf{0.865} & 0.471 & 0.777 & 0.720 \\
& Structured + ClinicalBERT & 0.800 & 0.909 & 0.851 & 0.559 & 0.787 & 0.734 \\
& Ours  & 0.887	& 0.833	& 0.859 	& 0.794	& 0.849	& 0.814 \\ \hline
\multirow{5}{*}{Cancer Cachexia} & ICD Codes &  0.780	& 0.557 &	0.650 &	0.593 & 0.754 & 0.575 \\
& Structured Data Only   & 0.722 &	1.00 & 	0.838 & 0 & 0.722 & 0.500 \\
& Structured + NCR & 1.000 & 0.211 & 0.348 & 1.000 & 0.661 & 0.605 \\
& Structured + ClinicalBERT & 0.857 & 0.105 & 0.188 & 0.977 & 0.600 & 0.541 \\
& Ours & 1.00 &	0.757 &	\textbf{0.862} & 1.00 & 0.932 & \textbf{0.878} \\ \hline
\multirow{5}{*}{Lupus Nephritis} & ICD Codes   & 0.979 & 0.758 & 0.855 & 0.974 & 0.892 & 0.866 \\
& Structured Data Only   & 0.933 &	0.677 &	0.785 &	0.921 &	0.832 &	0.799 \\
& Structured + NCR & 0.981 & 0.855 & 0.914 & 0.974 & 0.929 & 0.914 \\
& Structured + ClinicalBERT  & 0.919 & 0.919 & 0.919 & 0.868 & 0.895 & 0.894 \\
& Ours & 0.967 & 0.952 & \textbf{0.959} & 0.947 & 0.950 & \textbf{0.949}  \\ \hline
\end{tabular}%
\caption{Comparison of the disease classifier created by the proposed workflow with baseline methods on the gold testing subset. The method with the highest F1 score and AUC-ROC for each disease is highlighted in bold. Please note all methods except ICD Codes use our AutoML to build disease classifiers. }
\label{tab:disease_results}
\end{table}

Table \ref{tab:disease_results} compares the proposed workflow with the baseline methods. While the ICD codes tend to have high precision except for Cancer Cachexia, the proposed workflow achieves significantly higher recall than the ICD codes for Ovarian Cancer (0.962 vs 0.714), Lung Cancer (0.833 vs 0.727), Cancer Cachexia (0.757 vs 0.557), and Lupus Nephritis (0.952 vs 0.758), which suggests the disease classifier is more sensitive and tends to find more positive patients. This also leads to a higher F1 score of the disease classifiers for Ovarian Cancer (0.901 vs 0.814), Lung Cancer (0.859 vs 0.828), Cancer Cachexia (0.862 vs 0.650) and Lupus Nephritis (0.959 vs 0.855). In addition, the benefit of leveraging phenotypic features from unstructured notes by phenotyping NLP algorithms is also observed as the proposed workflow achieves significantly higher F1, AUC-PR and AUC-ROC than the baseline method which uses structured data only across all four diseases. Moreover, the phenotyping NLP algorithm \cite{zhang-etal-2021-self} used by the proposed workflow outperforms other phenotyping NLP algorithms, namely NCR and fine-tuned ClinicalBERT, for patient identification with consistently better AUC-ROC across four diseases.

\begin{table}[h]
\centering
\begin{tabular}{@{}c|c|ccc@{}}
\toprule
\textbf{Disease Name} & \textbf{Method} & \textbf{\makecell{Number of Predicted \\ Positive EHRs $N_{\text{pred}}$}} & \textbf{\makecell{Estimated \\ Precision $P_{\text{est}}$}} & \textbf{\makecell{Estimated Number of Actual \\ Positive EHRs $N_{\text{pred}} \times P_{\text{est}}$}}  \\ \midrule 
\multirow{2}{*}{Ovarian Cancer} & ICD Codes  & 38 & 0.946 & 36  \\
& Ours  & 143 & 0.776 & 111 \\ \hline
\multirow{2}{*}{Lung Cancer} & ICD Codes  &  585 & 0.960 &  562 \\
& Ours  & 1209 & 0.766  & 926 \\ \hline
\multirow{2}{*}{Cachexia in Cancer} & ICD Codes  & 51 & 0.780 & 40   \\
& Ours  & 326 & 0.969 & 316 \\ \hline
\multirow{2}{*}{Lupus Nephritis} & ICD Codes  & 63 & 0.979 &  62 \\
& Ours  & 142 &  0.758 & 108\\ \hline
\end{tabular}%
\caption{Comparison of the disease classifier created by the proposed workflow with baseline methods on the entire testing set. Please note $P_{est}$ of ICD Codes is taken from Table \ref{tab:disease_results} while $P_{est}$ of our proposed workflow is computed following the second evaluation method in section \ref{sec:evaluation_method}.}
\label{tab:disease_large_results}
\end{table}

Table \ref{tab:disease_large_results} evaluates the proposed workflow on the entire testing set which has 29,339 EHRs. The estimated precision of the proposed workflow is computed based on validation from clinicians while the estimated precision of ICD codes is taken from Table \ref{tab:disease_results}. For Cancer Cachexia, the proposed workflow achieves higher estimated precision and a larger number of positive predictions than the ICD codes. This leads to a significantly higher estimation of actual positive EHRs founded by the proposed workflow than the ICD codes (316 vs 40) which is 690\% more EHRs. For Ovarian Cancer, Lung Cancer and Lupus Nephritis, the proposed workflow has lower estimated precision but predicts much more EHRs as positive than ICD codes, which overall lead to finding more actual positive EHRs.

\section{Discussion}


This work conceptualises a machine-learning based approach to screen patient records automatically to identify miscoded, undiagnosed or at-risk patients at scale across multiple diseases with minimal technical and clinical input required for configuration and validation. Subject to validation by clinicians, the identified patients can be recruited to conduct epidemiological investigations, clinical trials, new treatment, and early interventions which can potentially improve patient outcomes. The section further discusses the insights of this study.


\subsection{Feature Analysis}

The contribution of critical features to the decision-making of the machine-learning disease classifiers for the four diseases is measured by SHAP values \cite{SHAP}. SHAP value is designed based on Shapley values to explain black-box models by quantifying the importance of an input feature and whether it impacts positively or negatively the outcome. Intuitively, the impact of a feature is measured by the variation of the model output when a given feature is observed versus when it is discarded.




\begin{figure}[!h]
\centering
\includegraphics[width=0.75\linewidth]{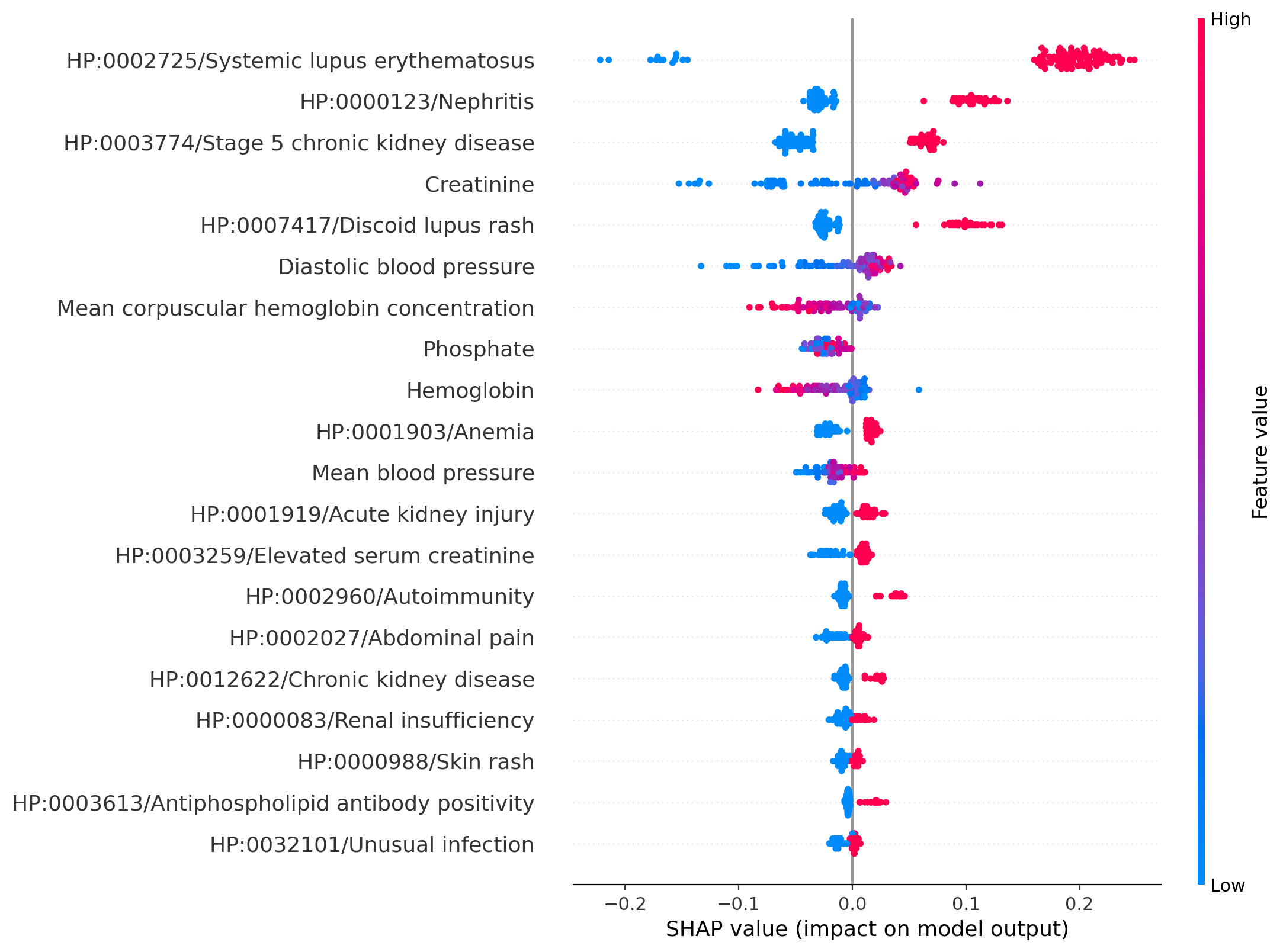}
\caption{A beeswarm plot of SHAP values highlighting the key features for the Lupus Nephritis classifier. Each dot on this plot represents one patient with blue corresponding to lower values or absence and red to higher ones or presence of clinical features.}
\label{fig:LN_shap}
\end{figure}

Figure \ref{fig:LN_shap} highlights the top features for the Lupus Nephritis classifier based on SHAP values. The classifier identifies several key markers of lupus such as systemic lupus erythematosus (SLE), discoid lupus rash and skin rash, as well as kidney diseases like nephritis. These affirm that the proposed workflow with AutoML (which selects top features automatically) and Clinician-in-the-Loop (which validates clinically relevant features) can build disease classifiers with interpretability and clinical validity. Similarity, the Ovarian Cancer classifier finds ovarian neoplasm, ascites, pelvic mass and more as top features, the Lung Cancer classifier finds the neoplasm of the lung, small cell lung carcinoma and more as top features, and the Cancer Cachexia classifier finds cachexia, neoplasm, poor appetite, weight loss and more as top features.

\begin{figure}[!h]
\centering
\includegraphics[width=0.9\linewidth]{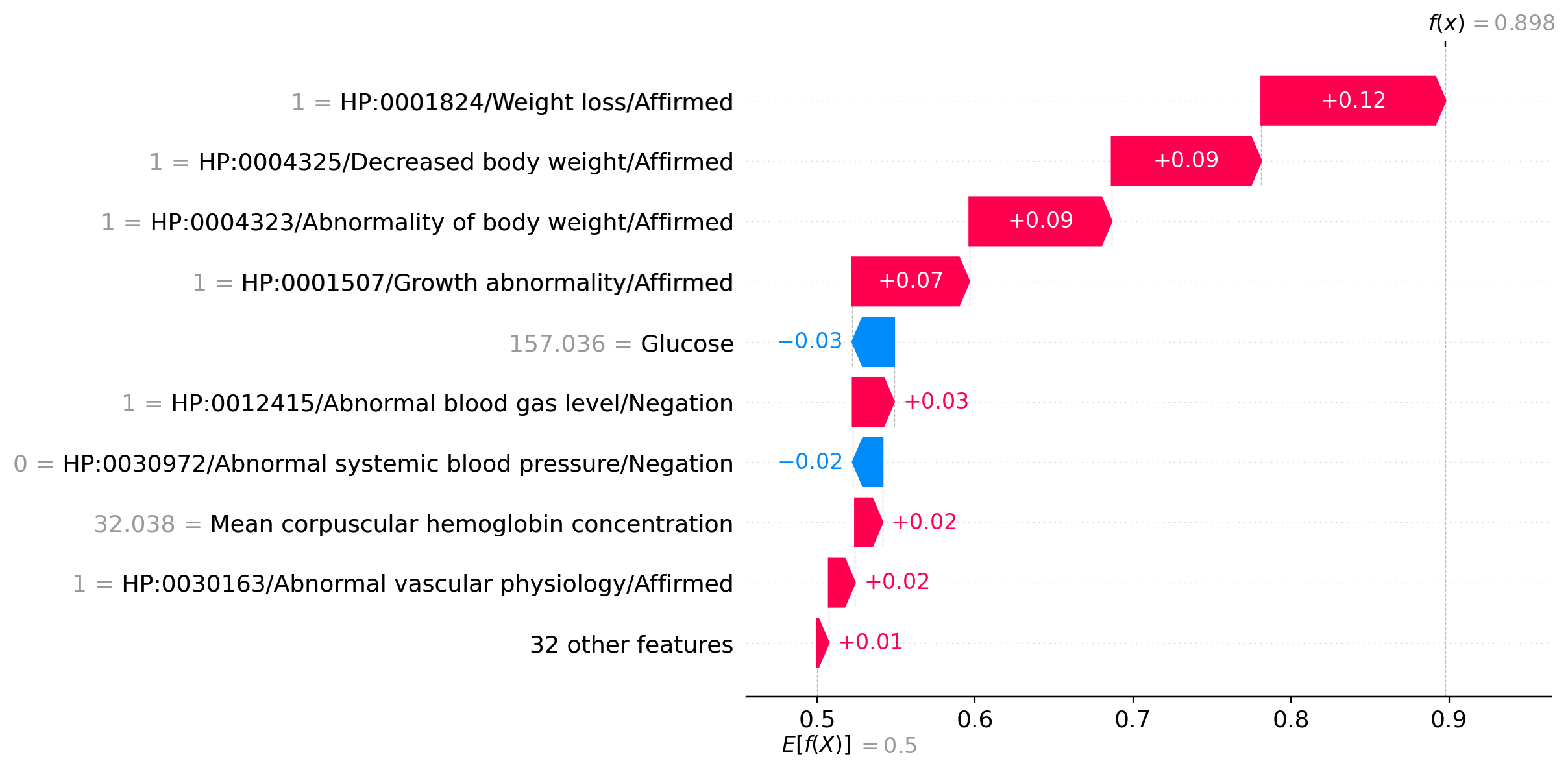}
\caption{A plot showing the impact of top features on the decision making for one patient. As shown, the presence of weight loss and its related phenotypes and neoplasm for this patient drastically increases the risk of Cancer Cachexia. The red bar represents the presence or higher values of a feature and blue represents absence or lower values. }
\label{fig:cachexia_ind_shap}
\end{figure}

Figure \ref{fig:cachexia_ind_shap} demonstrates the impact of top features on the decision making for one individual patient based on SHAP values. As the patient is recorded to have weight loss with the background condition of cancer, the Cancer Cachexia classifier predicts a high risk for the patient. Similarly, the decision making behind other disease classifiers for each patient can also be interpreted by SHAP values.




\subsection{The Benefits of Leveraging Phenotypes from Unstructured Textual Data}

Our results highlight the benefits of leveraging the phenotypic features from unstructured notes in identifying patients with diseases. Using clinical features from structured data alone results in poor performance as shown in Table \ref{tab:disease_results}, likely due to a large number of missing values (for example, only 36\% of patients had both height and weight data), mistakes and inconsistency of measurement units. 
For example, some patients have body weights and temperatures negative or in the millions. The units are also inconsistent between kilograms to pound and Celcius to Farenheit for body weight and temperature, respectively. Although the data preprocessing steps clean the dataset to impute remove extreme values and unify units, this suggests potential inconsistencies and mistakes present in the structured fields. Thus, there is great value in extracting the phenotypes directly from the source clinical text itself. As the results above show, utilising the unstructured notes leads to an improved sensitivity which is critical for finding patients with rare conditions at scale. 

Another benefit of extracting phenotypes from clinical text is that it allows for patients to be identified based on a known phenotypic profile, instead of simply querying for patients based on ICD codes. In other words, the disease classifier in Figure \ref{fig:workflow} can be constructed by a number of inclusive and exclusive pre-defined criteria based on phenotypes rather than AutoML. For example, the work \cite{yu2018enabling} identifies patients with coronary artery disease, rheumatoid arthritis, Crohn's disease, and ulcerative colitis based on four specific phenotypes, respectively.
Though such a method requires prior clinical knowledge of diseases and explicit phenotype mentions in unstructured notes for all inclusive and exclusive criteria, the method does not require any retrospective labelled data for training and can be applied directly to a prospective cohort to identify suitable patients. However, for diseases whose diagnosis criteria may not be clear yet, AutoML can be useful to provide statistical insights on phenotypic profiles of positive patient cohorts to understand the diseases better.

\subsection{The Benefits of Incorporating Gold Labels into Training via Clinician-in-the-Loop}

Our approach also underscores the benefits of incorporating gold standard labels via Clinician-in-the-Loop. Since it is not feasible to create manually validated large-scale datasets to train classifiers, the proposed workflow minimises the required input from experts, producing improved classification performance. Such an approach can be deployed on both prospective and retrospective data depending on the use cases. Given the multiple flaws of common medical coding systems (such as ICD) highlighted in this paper, the proposed workflow is highly suited for the identification of suitable patients for clinical trials from a patient dataset, finding more patients than otherwise found by querying medical codes. Such classifiers (once trained) can also be deployed as a clinical decision support tool to alert clinicians of patients potentially at risk such that they can be taken through appropriate screening and treatment pathways earlier, with a direct improvement in patient experience and outcomes. 

\subsection{Future Work}

The proposed workflow can be applied to other, general hospital admission datasets for not only the identification of patients with target diseases, but also for the early prediction of diseases by training these classifiers on retrospective longitudinal data. The advantage of the proposed workflow is that it is designed to be use-case and disease agnostic.

This work could be further extended by extracting demographic variables (such as age, gender, ethnicity, dates) and social factor variables (such as smoking habit, and alcohol consumption) directly from the clinical notes themselves. Due to the limitations of the MIMIC-III dataset used (which adds significant noise to some demographic variables to preserve anonymity), there is little value in extracting such variables for this study. However, the critical next step for this work is to apply the approach to more real-world data obtained from hospital admissions and conduct a prospective evaluation, using a greater number of patients for testing and evaluation. 

Moreover, some components in the proposed workflow can be further explored by using alternative approaches. For example, the disease classifiers can be deep neural networks or reinforcement learning models; the clinician-in-the-loop mechanism may consider using active learning strategies to reduce the number of gold labels required for training. 
\section{Conclusions}

This work thus presents a workflow to build machine learning classifiers to identify patients with certain diseases from EHRs. There is expert-involvement via Clinician-in-the-Loop of this workflow and the  feedback process helps the machine learning classifiers evolve with clinical feedback about relevant clinical features and gold diagnosis labels. The performance of the proposed workflow has been demonstrated to be superior to the ICD-code baseline, with the machine-learning based approach finding significantly more patients than if only ICD codes are used. 

This work also demonstrates the importance of extracting phenotypic information from unstructured textual data. Structured clinical features alone have poor performance (when used in a classifier), due to missing and often unreliable values. Extracting phenotypes from clinical text allows for significantly more information to be retrieved from EHRs which results in improved classification accuracy and better interpretation of top impact features. 

The approach presented in this work is designed to be disease and use-case agnostic, and can potentially be applied at scale across multiple diseases in future works. The Clinician-in-the-Loop and AutoML framework lead to rapid configuration of models, minimising clinicians' and engineers' inputs, while maintaining good accuracy and clinical validity.
\section{Ethics Considerations}
The research has been carried out in accordance with relevant guidelines and regulations for the MIMIC-III data. Clinical experts support us annotating data and validating model predictions and relevant features. The ultimate objective of the proposed method is to identify patients at high risk of particular diseases earlier and at scale to assist clinicians in their decisions. Before the deployment of the proposed method in the actual clinical setting, the proposed method is subject to systematic debugging, extensive simulation, testing and validation under the supervision of expert clinicians following related regulatory guidelines.

\section{Code and Data Availability}
The MIMIC-III dataset is publicly available at \url{https://mimic.mit.edu/} (accessed in November 2021) and the code for preprocessing is published \cite{harutyunyan2019multitask}. The source code of the workflow and gold labels can not be shared publicly due to their proprietary nature, however, the methodology and the annotation guideline of the gold labels are described in Section \ref{sec:methodology}.

\section{Competing Interests}
This work is under collaboration with  Imperial College London, Hong Kong Baptist University and Pangaea Data Limited.

\section{Author Contributions}
JZ: Conceptualization, Methodology, Development, Resources, Writing Manuscript.
AS: Conceptualization, Methodology, Experiments, Writing Manuscript.
LB: Conceptualization, Methodology, Experiments, Development.
TL: Conceptualization, Experiments, Development.
AT: Conceptualization, Experiments.
VG: Conceptualization, Review, Supervision, Project Administration.
YG: Conceptualization, Review, Supervision, Project Administration.

\section{Acknowledgements}
We would like to thank Dr. Garima Gupta, Dr. Deepa (M.R.S.H) and Dr. Ashok (M.S.) for helping us create gold-standard phenotype annotation data, create gold-standard labels of disease diagnosis and validate relevant clinical features of diseases. 

\bibliography{reference,anthology}


\begin{thebibliography}{49}
\ifx \bisbn   \undefined \def \bisbn  #1{ISBN #1}\fi
\ifx \binits  \undefined \def \binits#1{#1}\fi
\ifx \bauthor  \undefined \def \bauthor#1{#1}\fi
\ifx \batitle  \undefined \def \batitle#1{#1}\fi
\ifx \bjtitle  \undefined \def \bjtitle#1{#1}\fi
\ifx \bvolume  \undefined \def \bvolume#1{\textbf{#1}}\fi
\ifx \byear  \undefined \def \byear#1{#1}\fi
\ifx \bissue  \undefined \def \bissue#1{#1}\fi
\ifx \bfpage  \undefined \def \bfpage#1{#1}\fi
\ifx \blpage  \undefined \def \blpage #1{#1}\fi
\ifx \burl  \undefined \def \burl#1{\textsf{#1}}\fi
\ifx \doiurl  \undefined \def \doiurl#1{\textsf{#1}}\fi
\ifx \betal  \undefined \def \betal{\textit{et al.}}\fi
\ifx \binstitute  \undefined \def \binstitute#1{#1}\fi
\ifx \binstitutionaled  \undefined \def \binstitutionaled#1{#1}\fi
\ifx \bctitle  \undefined \def \bctitle#1{#1}\fi
\ifx \beditor  \undefined \def \beditor#1{#1}\fi
\ifx \bpublisher  \undefined \def \bpublisher#1{#1}\fi
\ifx \bbtitle  \undefined \def \bbtitle#1{#1}\fi
\ifx \bedition  \undefined \def \bedition#1{#1}\fi
\ifx \bseriesno  \undefined \def \bseriesno#1{#1}\fi
\ifx \blocation  \undefined \def \blocation#1{#1}\fi
\ifx \bsertitle  \undefined \def \bsertitle#1{#1}\fi
\ifx \bsnm \undefined \def \bsnm#1{#1}\fi
\ifx \bsuffix \undefined \def \bsuffix#1{#1}\fi
\ifx \bparticle \undefined \def \bparticle#1{#1}\fi
\ifx \barticle \undefined \def \barticle#1{#1}\fi
\ifx \bconfdate \undefined \def \bconfdate #1{#1}\fi
\ifx \botherref \undefined \def \botherref #1{#1}\fi
\ifx \url \undefined \def \url#1{\textsf{#1}}\fi
\ifx \bchapter \undefined \def \bchapter#1{#1}\fi
\ifx \bbook \undefined \def \bbook#1{#1}\fi
\ifx \bcomment \undefined \def \bcomment#1{#1}\fi
\ifx \oauthor \undefined \def \oauthor#1{#1}\fi
\ifx \citeauthoryear \undefined \def \citeauthoryear#1{#1}\fi
\ifx \endbibitem  \undefined \def \endbibitem {}\fi
\ifx \bconflocation  \undefined \def \bconflocation#1{#1}\fi
\ifx \arxivurl  \undefined \def \arxivurl#1{\textsf{#1}}\fi
\csname PreBibitemsHook\endcsname

\bibitem{NHSActivity}
\begin{botherref}
Activity in the NHS
(2020).
\url{https://www.kingsfund.org.uk/projects/nhs-in-a-nutshell/NHS-activity}
\end{botherref}
\endbibitem

\bibitem{Dash2019}
\begin{botherref}
\oauthor{\bsnm{Dash}, \binits{S.}},
\oauthor{\bsnm{Shakyawar}, \binits{S.K.}},
\oauthor{\bsnm{Sharma}, \binits{M.}},
\oauthor{\bsnm{Kaushik}, \binits{S.}}:
Big data in healthcare: management, analysis and future prospects.
Journal of Big Data
\textbf{6}(1)
(2019).
doi:\doiurl{10.1186/s40537-019-0217-0}
\end{botherref}
\endbibitem

\bibitem{Vamathevan2019}
\begin{barticle}
\bauthor{\bsnm{Vamathevan}, \binits{J.}},
\bauthor{\bsnm{Clark}, \binits{D.}},
\bauthor{\bsnm{Czodrowski}, \binits{P.}},
\bauthor{\bsnm{Dunham}, \binits{I.}},
\bauthor{\bsnm{Ferran}, \binits{E.}},
\bauthor{\bsnm{Lee}, \binits{G.}},
\bauthor{\bsnm{Li}, \binits{B.}},
\bauthor{\bsnm{Madabhushi}, \binits{A.}},
\bauthor{\bsnm{Shah}, \binits{P.}},
\bauthor{\bsnm{Spitzer}, \binits{M.}},
\bauthor{\bsnm{Zhao}, \binits{S.}}:
\batitle{Applications of machine learning in drug discovery and development}.
\bjtitle{Nature Reviews Drug Discovery}
\bvolume{18}(\bissue{6}),
\bfpage{463}--\blpage{477}
(\byear{2019}).
doi:\doiurl{10.1038/s41573-019-0024-5}
\end{barticle}
\endbibitem

\bibitem{Park2021}
\begin{botherref}
\oauthor{\bsnm{Park}, \binits{D.J.}},
\oauthor{\bsnm{Park}, \binits{M.W.}},
\oauthor{\bsnm{Lee}, \binits{H.}},
\oauthor{\bsnm{Kim}, \binits{Y.-J.}},
\oauthor{\bsnm{Kim}, \binits{Y.}},
\oauthor{\bsnm{Park}, \binits{Y.H.}}:
Development of machine learning model for diagnostic disease prediction based
  on laboratory tests.
Scientific Reports
\textbf{11}(1)
(2021).
doi:\doiurl{10.1038/s41598-021-87171-5}
\end{botherref}
\endbibitem

\bibitem{Hassaine2020}
\begin{barticle}
\bauthor{\bsnm{Hassaine}, \binits{A.}},
\bauthor{\bsnm{Salimi-Khorshidi}, \binits{G.}},
\bauthor{\bsnm{Canoy}, \binits{D.}},
\bauthor{\bsnm{Rahimi}, \binits{K.}}:
\batitle{Untangling the complexity of multimorbidity with machine learning}.
\bjtitle{Mechanisms of Ageing and Development}
\bvolume{190},
\bfpage{111325}
(\byear{2020}).
doi:\doiurl{10.1016/j.mad.2020.111325}
\end{barticle}
\endbibitem

\bibitem{Wockenfuss2009}
\begin{barticle}
\bauthor{\bsnm{Wockenfuss}, \binits{R.}},
\bauthor{\bsnm{Frese}, \binits{T.}},
\bauthor{\bsnm{Herrmann}, \binits{K.}},
\bauthor{\bsnm{Claussnitzer}, \binits{M.}},
\bauthor{\bsnm{Sandholzer}, \binits{H.}}:
\batitle{Three- and four-digit {ICD}-10 is not a reliable classification system
  in primary care}.
\bjtitle{Scandinavian Journal of Primary Health Care}
\bvolume{27}(\bissue{3}),
\bfpage{131}--\blpage{136}
(\byear{2009}).
doi:\doiurl{10.1080/02813430903072215}
\end{barticle}
\endbibitem

\bibitem{southern_etal}
\begin{barticle}
\bauthor{\bsnm{Southern}, \binits{D.A.}},
\bauthor{\bsnm{Hall}, \binits{M.}},
\bauthor{\bsnm{White}, \binits{D.E.}},
\bauthor{\bsnm{Romano}, \binits{P.S.}},
\bauthor{\bsnm{Sundararajan}, \binits{V.}},
\bauthor{\bsnm{Droesler}, \binits{S.E.}},
\bauthor{\bsnm{Pincus}, \binits{H.A.}},
\bauthor{\bsnm{Ghali}, \binits{W.A.}}:
\batitle{{Opportunities and challenges for quality and safety applications in
  ICD-11: an international survey of users of coded health data}}.
\bjtitle{International Journal for Quality in Health Care}
\bvolume{28}(\bissue{1}),
\bfpage{129}--\blpage{135}
(\byear{2015}).
doi:\doiurl{10.1093/intqhc/mzv096}.
\arxivurl{https://academic.oup.com/intqhc/article-pdf/28/1/129/6991704/mzv096.pdf}
\end{barticle}
\endbibitem

\bibitem{Snyder2017}
\begin{barticle}
\bauthor{\bsnm{Snyder}, \binits{A.B.}},
\bauthor{\bsnm{Lane}, \binits{P.A.}},
\bauthor{\bsnm{Zhou}, \binits{M.}},
\bauthor{\bsnm{Paulukonis}, \binits{S.T.}},
\bauthor{\bsnm{Hulihan}, \binits{M.M.}}:
\batitle{The accuracy of hospital icd-9-cm codes for determining sickle cell
  disease genotype}.
\bjtitle{Journal of rare diseases research {\&} treatment}
\bvolume{2}(\bissue{4}),
\bfpage{39}--\blpage{45}
(\byear{2017}).
\bcomment{29202133[pmid]}
\end{barticle}
\endbibitem

\bibitem{Horsky2018}
\begin{barticle}
\bauthor{\bsnm{Horsky}, \binits{J.}},
\bauthor{\bsnm{Drucker}, \binits{E.A.}},
\bauthor{\bsnm{Ramelson}, \binits{H.Z.}}:
\batitle{Accuracy and completeness of clinical coding using icd-10 for
  ambulatory visits}.
\bjtitle{AMIA ... Annual Symposium proceedings. AMIA Symposium}
\bvolume{2017},
\bfpage{912}--\blpage{920}
(\byear{2018}).
\bcomment{29854158[pmid]}
\end{barticle}
\endbibitem

\bibitem{Burles2017}
\begin{botherref}
\oauthor{\bsnm{Burles}, \binits{K.}},
\oauthor{\bsnm{Innes}, \binits{G.}},
\oauthor{\bsnm{Senior}, \binits{K.}},
\oauthor{\bsnm{Lang}, \binits{E.}},
\oauthor{\bsnm{McRae}, \binits{A.}}:
Limitations of pulmonary embolism {ICD}-10 codes in emergency department
  administrative data: let the buyer beware.
{BMC} Medical Research Methodology
\textbf{17}(1)
(2017).
doi:\doiurl{10.1186/s12874-017-0361-1}
\end{botherref}
\endbibitem

\bibitem{xu2020identifying}
\begin{barticle}
\bauthor{\bsnm{Xu}, \binits{Z.}},
\bauthor{\bsnm{Chou}, \binits{J.}},
\bauthor{\bsnm{Zhang}, \binits{X.S.}},
\bauthor{\bsnm{Luo}, \binits{Y.}},
\bauthor{\bsnm{Isakova}, \binits{T.}},
\bauthor{\bsnm{Adekkanattu}, \binits{P.}},
\bauthor{\bsnm{Ancker}, \binits{J.S.}},
\bauthor{\bsnm{Jiang}, \binits{G.}},
\bauthor{\bsnm{Kiefer}, \binits{R.C.}},
\bauthor{\bsnm{Pacheco}, \binits{J.A.}}, \betal:
\batitle{Identifying sub-phenotypes of acute kidney injury using structured and
  unstructured electronic health record data with memory networks}.
\bjtitle{Journal of biomedical informatics}
\bvolume{102},
\bfpage{103361}
(\byear{2020})
\end{barticle}
\endbibitem

\bibitem{liu2021evaluating}
\begin{barticle}
\bauthor{\bsnm{Liu}, \binits{R.}},
\bauthor{\bsnm{Rizzo}, \binits{S.}},
\bauthor{\bsnm{Whipple}, \binits{S.}},
\bauthor{\bsnm{Pal}, \binits{N.}},
\bauthor{\bsnm{Pineda}, \binits{A.L.}},
\bauthor{\bsnm{Lu}, \binits{M.}},
\bauthor{\bsnm{Arnieri}, \binits{B.}},
\bauthor{\bsnm{Lu}, \binits{Y.}},
\bauthor{\bsnm{Capra}, \binits{W.}},
\bauthor{\bsnm{Copping}, \binits{R.}}, \betal:
\batitle{Evaluating eligibility criteria of oncology trials using real-world
  data and ai}.
\bjtitle{Nature}
\bvolume{592}(\bissue{7855}),
\bfpage{629}--\blpage{633}
(\byear{2021})
\end{barticle}
\endbibitem

\bibitem{wu2019prediction}
\begin{barticle}
\bauthor{\bsnm{Wu}, \binits{C.-C.}},
\bauthor{\bsnm{Yeh}, \binits{W.-C.}},
\bauthor{\bsnm{Hsu}, \binits{W.-D.}},
\bauthor{\bsnm{Islam}, \binits{M.M.}},
\bauthor{\bsnm{Nguyen}, \binits{P.A.A.}},
\bauthor{\bsnm{Poly}, \binits{T.N.}},
\bauthor{\bsnm{Wang}, \binits{Y.-C.}},
\bauthor{\bsnm{Yang}, \binits{H.-C.}},
\bauthor{\bsnm{Li}, \binits{Y.-C.J.}}:
\batitle{Prediction of fatty liver disease using machine learning algorithms}.
\bjtitle{Computer methods and programs in biomedicine}
\bvolume{170},
\bfpage{23}--\blpage{29}
(\byear{2019})
\end{barticle}
\endbibitem

\bibitem{sekelj2021detecting}
\begin{barticle}
\bauthor{\bsnm{Sekelj}, \binits{S.}},
\bauthor{\bsnm{Sandler}, \binits{B.}},
\bauthor{\bsnm{Johnston}, \binits{E.}},
\bauthor{\bsnm{Pollock}, \binits{K.G.}},
\bauthor{\bsnm{Hill}, \binits{N.R.}},
\bauthor{\bsnm{Gordon}, \binits{J.}},
\bauthor{\bsnm{Tsang}, \binits{C.}},
\bauthor{\bsnm{Khan}, \binits{S.}},
\bauthor{\bsnm{Ng}, \binits{F.S.}},
\bauthor{\bsnm{Farooqui}, \binits{U.}}:
\batitle{Detecting undiagnosed atrial fibrillation in uk primary care:
  validation of a machine learning prediction algorithm in a retrospective
  cohort study}.
\bjtitle{European journal of preventive cardiology}
\bvolume{28}(\bissue{6}),
\bfpage{598}--\blpage{605}
(\byear{2021})
\end{barticle}
\endbibitem

\bibitem{Chen2021}
\begin{barticle}
\bauthor{\bsnm{Chen}, \binits{Y.}},
\bauthor{\bsnm{Huang}, \binits{S.}},
\bauthor{\bsnm{Chen}, \binits{T.}},
\bauthor{\bsnm{Liang}, \binits{D.}},
\bauthor{\bsnm{Yang}, \binits{J.}},
\bauthor{\bsnm{Zeng}, \binits{C.}},
\bauthor{\bsnm{Li}, \binits{X.}},
\bauthor{\bsnm{Xie}, \binits{G.}},
\bauthor{\bsnm{Liu}, \binits{Z.}}:
\batitle{Machine learning for prediction and risk stratification of lupus
  nephritis renal flare}.
\bjtitle{American Journal of Nephrology}
\bvolume{52}(\bissue{2}),
\bfpage{152}--\blpage{160}
(\byear{2021}).
doi:\doiurl{10.1159/000513566}
\end{barticle}
\endbibitem

\bibitem{Kegerreis2019}
\begin{botherref}
\oauthor{\bsnm{Kegerreis}, \binits{B.}},
\oauthor{\bsnm{Catalina}, \binits{M.D.}},
\oauthor{\bsnm{Bachali}, \binits{P.}},
\oauthor{\bsnm{Geraci}, \binits{N.S.}},
\oauthor{\bsnm{Labonte}, \binits{A.C.}},
\oauthor{\bsnm{Zeng}, \binits{C.}},
\oauthor{\bsnm{Stearrett}, \binits{N.}},
\oauthor{\bsnm{Crandall}, \binits{K.A.}},
\oauthor{\bsnm{Lipsky}, \binits{P.E.}},
\oauthor{\bsnm{Grammer}, \binits{A.C.}}:
Machine learning approaches to predict lupus disease activity from gene
  expression data.
Scientific Reports
\textbf{9}(1)
(2019).
doi:\doiurl{10.1038/s41598-019-45989-0}
\end{botherref}
\endbibitem

\bibitem{Wei2015}
\begin{barticle}
\bauthor{\bsnm{Wei}, \binits{W.-Q.}},
\bauthor{\bsnm{Teixeira}, \binits{P.L.}},
\bauthor{\bsnm{Mo}, \binits{H.}},
\bauthor{\bsnm{Cronin}, \binits{R.M.}},
\bauthor{\bsnm{Warner}, \binits{J.L.}},
\bauthor{\bsnm{Denny}, \binits{J.C.}}:
\batitle{Combining billing codes, clinical notes, and medications from
  electronic health records provides superior phenotyping performance}.
\bjtitle{Journal of the American Medical Informatics Association}
\bvolume{23}(\bissue{e1}),
\bfpage{20}--\blpage{27}
(\byear{2015}).
doi:\doiurl{10.1093/jamia/ocv130}
\end{barticle}
\endbibitem

\bibitem{Kong2019}
\begin{barticle}
\bauthor{\bsnm{Kong}, \binits{H.-J.}}:
\batitle{Managing unstructured big data in healthcare system}.
\bjtitle{Healthcare Informatics Research}
\bvolume{25}(\bissue{1}),
\bfpage{1}
(\byear{2019}).
doi:\doiurl{10.4258/hir.2019.25.1.1}
\end{barticle}
\endbibitem

\bibitem{xu2019multimodal}
\begin{bchapter}
\bauthor{\bsnm{Xu}, \binits{K.}},
\bauthor{\bsnm{Lam}, \binits{M.}},
\bauthor{\bsnm{Pang}, \binits{J.}},
\bauthor{\bsnm{Gao}, \binits{X.}},
\bauthor{\bsnm{Band}, \binits{C.}},
\bauthor{\bsnm{Mathur}, \binits{P.}},
\bauthor{\bsnm{Papay}, \binits{F.}},
\bauthor{\bsnm{Khanna}, \binits{A.K.}},
\bauthor{\bsnm{Cywinski}, \binits{J.B.}},
\bauthor{\bsnm{Maheshwari}, \binits{K.}}, \betal:
\bctitle{Multimodal machine learning for automated icd coding}.
In: \bbtitle{Machine Learning for Healthcare Conference},
pp. \bfpage{197}--\blpage{215}
(\byear{2019}).
\bcomment{PMLR}
\end{bchapter}
\endbibitem

\bibitem{zhang-etal-2020-bert}
\begin{bchapter}
\bauthor{\bsnm{Zhang}, \binits{Z.}},
\bauthor{\bsnm{Liu}, \binits{J.}},
\bauthor{\bsnm{Razavian}, \binits{N.}}:
\bctitle{{BERT}-{XML}: Large scale automated {ICD} coding using {BERT}
  pretraining}.
In: \bbtitle{Proceedings of the 3rd Clinical Natural Language Processing
  Workshop},
pp. \bfpage{24}--\blpage{34}.
\bpublisher{Association for Computational Linguistics},
\blocation{Online}
(\byear{2020}).
doi:\doiurl{10.18653/v1/2020.clinicalnlp-1.3}.
\burl{https://aclanthology.org/2020.clinicalnlp-1.3}
\end{bchapter}
\endbibitem

\bibitem{robinson2012deep}
\begin{barticle}
\bauthor{\bsnm{Robinson}, \binits{P.N.}}:
\batitle{Deep phenotyping for precision medicine}.
\bjtitle{Human mutation}
\bvolume{33}(\bissue{5}),
\bfpage{777}--\blpage{780}
(\byear{2012})
\end{barticle}
\endbibitem

\bibitem{yu2018enabling}
\begin{barticle}
\bauthor{\bsnm{Yu}, \binits{S.}},
\bauthor{\bsnm{Ma}, \binits{Y.}},
\bauthor{\bsnm{Gronsbell}, \binits{J.}},
\bauthor{\bsnm{Cai}, \binits{T.}},
\bauthor{\bsnm{Ananthakrishnan}, \binits{A.N.}},
\bauthor{\bsnm{Gainer}, \binits{V.S.}},
\bauthor{\bsnm{Churchill}, \binits{S.E.}},
\bauthor{\bsnm{Szolovits}, \binits{P.}},
\bauthor{\bsnm{Murphy}, \binits{S.N.}},
\bauthor{\bsnm{Kohane}, \binits{I.S.}}, \betal:
\batitle{Enabling phenotypic big data with phenorm}.
\bjtitle{Journal of the American Medical Informatics Association}
\bvolume{25}(\bissue{1}),
\bfpage{54}--\blpage{60}
(\byear{2018})
\end{barticle}
\endbibitem

\bibitem{SLE_Criteria}
\begin{barticle}
\bauthor{\bsnm{Aringer}, \binits{M.}},
\bauthor{\bsnm{Costenbader}, \binits{K.}},
\bauthor{\bsnm{Daikh}, \binits{D.}},
\bauthor{\bsnm{Brinks}, \binits{R.}},
\bauthor{\bsnm{Mosca}, \binits{M.}},
\bauthor{\bsnm{Ramsey-Goldman}, \binits{R.}},
\bauthor{\bsnm{Smolen}, \binits{J.S.}},
\bauthor{\bsnm{Wofsy}, \binits{D.}},
\bauthor{\bsnm{Boumpas}, \binits{D.T.}},
\bauthor{\bsnm{Kamen}, \binits{D.L.}},
\bauthor{\bsnm{Jayne}, \binits{D.}},
\bauthor{\bsnm{Cervera}, \binits{R.}},
\bauthor{\bsnm{Costedoat-Chalumeau}, \binits{N.}},
\bauthor{\bsnm{Diamond}, \binits{B.}},
\bauthor{\bsnm{Gladman}, \binits{D.D.}},
\bauthor{\bsnm{Hahn}, \binits{B.}},
\bauthor{\bsnm{Hiepe}, \binits{F.}},
\bauthor{\bsnm{Jacobsen}, \binits{S.}},
\bauthor{\bsnm{Khanna}, \binits{D.}},
\bauthor{\bsnm{Lerstr{\o}m}, \binits{K.}},
\bauthor{\bsnm{Massarotti}, \binits{E.}},
\bauthor{\bsnm{McCune}, \binits{J.}},
\bauthor{\bsnm{Ruiz-Irastorza}, \binits{G.}},
\bauthor{\bsnm{Sanchez-Guerrero}, \binits{J.}},
\bauthor{\bsnm{Schneider}, \binits{M.}},
\bauthor{\bsnm{Urowitz}, \binits{M.}},
\bauthor{\bsnm{Bertsias}, \binits{G.}},
\bauthor{\bsnm{Hoyer}, \binits{B.F.}},
\bauthor{\bsnm{Leuchten}, \binits{N.}},
\bauthor{\bsnm{Tani}, \binits{C.}},
\bauthor{\bsnm{Tedeschi}, \binits{S.K.}},
\bauthor{\bsnm{Touma}, \binits{Z.}},
\bauthor{\bsnm{Schmajuk}, \binits{G.}},
\bauthor{\bsnm{Anic}, \binits{B.}},
\bauthor{\bsnm{Assan}, \binits{F.}},
\bauthor{\bsnm{Chan}, \binits{T.M.}},
\bauthor{\bsnm{Clarke}, \binits{A.E.}},
\bauthor{\bsnm{Crow}, \binits{M.K.}},
\bauthor{\bsnm{Czirj{\'a}k}, \binits{L.}},
\bauthor{\bsnm{Doria}, \binits{A.}},
\bauthor{\bsnm{Graninger}, \binits{W.}},
\bauthor{\bsnm{Halda-Kiss}, \binits{B.}},
\bauthor{\bsnm{Hasni}, \binits{S.}},
\bauthor{\bsnm{Izmirly}, \binits{P.M.}},
\bauthor{\bsnm{Jung}, \binits{M.}},
\bauthor{\bsnm{Kum{\'a}novics}, \binits{G.}},
\bauthor{\bsnm{Mariette}, \binits{X.}},
\bauthor{\bsnm{Padjen}, \binits{I.}},
\bauthor{\bsnm{Pego-Reigosa}, \binits{J.M.}},
\bauthor{\bsnm{Romero-Diaz}, \binits{J.}},
\bauthor{\bsnm{R{\'u}a-Figueroa~Fern{\'a}ndez}, \binits{{\'I}.}},
\bauthor{\bsnm{Seror}, \binits{R.}},
\bauthor{\bsnm{Stummvoll}, \binits{G.H.}},
\bauthor{\bsnm{Tanaka}, \binits{Y.}},
\bauthor{\bsnm{Tektonidou}, \binits{M.G.}},
\bauthor{\bsnm{Vasconcelos}, \binits{C.}},
\bauthor{\bsnm{Vital}, \binits{E.M.}},
\bauthor{\bsnm{Wallace}, \binits{D.J.}},
\bauthor{\bsnm{Yavuz}, \binits{S.}},
\bauthor{\bsnm{Meroni}, \binits{P.L.}},
\bauthor{\bsnm{Fritzler}, \binits{M.J.}},
\bauthor{\bsnm{Naden}, \binits{R.}},
\bauthor{\bsnm{D{\"o}rner}, \binits{T.}},
\bauthor{\bsnm{Johnson}, \binits{S.R.}}:
\batitle{2019 european league against rheumatism/american college of
  rheumatology classification criteria for systemic lupus erythematosus}.
\bjtitle{Arthritis {\&} rheumatology (Hoboken, N.J.)}
\bvolume{71}(\bissue{9}),
\bfpage{1400}--\blpage{1412}
(\byear{2019}).
doi:\doiurl{10.1002/art.40930}.
\bcomment{31385462[pmid]}
\end{barticle}
\endbibitem

\bibitem{deisseroth2019clinphen}
\begin{barticle}
\bauthor{\bsnm{Deisseroth}, \binits{C.A.}},
\bauthor{\bsnm{Birgmeier}, \binits{J.}},
\bauthor{\bsnm{Bodle}, \binits{E.E.}},
\bauthor{\bsnm{Kohler}, \binits{J.N.}},
\bauthor{\bsnm{Matalon}, \binits{D.R.}},
\bauthor{\bsnm{Nazarenko}, \binits{Y.}},
\bauthor{\bsnm{Genetti}, \binits{C.A.}},
\bauthor{\bsnm{Brownstein}, \binits{C.A.}},
\bauthor{\bsnm{Schmitz-Abe}, \binits{K.}},
\bauthor{\bsnm{Schoch}, \binits{K.}}, \betal:
\batitle{Clinphen extracts and prioritizes patient phenotypes directly from
  medical records to expedite genetic disease diagnosis}.
\bjtitle{Genetics in Medicine}
\bvolume{21}(\bissue{7}),
\bfpage{1585}--\blpage{1593}
(\byear{2019})
\end{barticle}
\endbibitem

\bibitem{jonquet2009ncbo}
\begin{bchapter}
\bauthor{\bsnm{Jonquet}, \binits{C.}},
\bauthor{\bsnm{Shah}, \binits{N.}},
\bauthor{\bsnm{Youn}, \binits{C.}},
\bauthor{\bsnm{Callendar}, \binits{C.}},
\bauthor{\bsnm{Storey}, \binits{M.-A.}},
\bauthor{\bsnm{Musen}, \binits{M.}}:
\bctitle{{NCBO Annotator: Semantic Annotation of Biomedical Data}}.
In: \bbtitle{International Semantic Web Conference, Poster and Demo Session},
vol. \bseriesno{110}
(\byear{2009})
\end{bchapter}
\endbibitem

\bibitem{Savova2010}
\begin{barticle}
\bauthor{\bsnm{Savova}, \binits{G.K.}},
\bauthor{\bsnm{Masanz}, \binits{J.J.}},
\bauthor{\bsnm{Ogren}, \binits{P.V.}},
\bauthor{\bsnm{Zheng}, \binits{J.}},
\bauthor{\bsnm{Sohn}, \binits{S.}},
\bauthor{\bsnm{Kipper-Schuler}, \binits{K.C.}},
\bauthor{\bsnm{Chute}, \binits{C.G.}}:
\batitle{{Mayo clinical Text Analysis and Knowledge Extraction System (cTAKES):
  architecture, component evaluation and applications}}.
\bjtitle{Journal of the American Medical Informatics Association : JAMIA}
\bvolume{17}(\bissue{5}),
\bfpage{507}--\blpage{513}
(\byear{2010}).
doi:\doiurl{10.1136/jamia.2009.001560}
\end{barticle}
\endbibitem

\bibitem{Aronson2010}
\begin{barticle}
\bauthor{\bsnm{Aronson}, \binits{A.R.}},
\bauthor{\bsnm{Lang}, \binits{F.-M.}}:
\batitle{{An overview of MetaMap: historical perspective and recent advances.}}
\bjtitle{Journal of the American Medical Informatics Association : JAMIA}
\bvolume{17}(\bissue{3}),
\bfpage{229}--\blpage{236}
(\byear{2010}).
doi:\doiurl{10.1136/jamia.2009.002733}
\end{barticle}
\endbibitem

\bibitem{zhang-etal-2021-self}
\begin{bchapter}
\bauthor{\bsnm{Zhang}, \binits{J.}},
\bauthor{\bsnm{Bolanos~Trujillo}, \binits{L.}},
\bauthor{\bsnm{Li}, \binits{T.}},
\bauthor{\bsnm{Tanwar}, \binits{A.}},
\bauthor{\bsnm{Freire}, \binits{G.}},
\bauthor{\bsnm{Yang}, \binits{X.}},
\bauthor{\bsnm{Ive}, \binits{J.}},
\bauthor{\bsnm{Gupta}, \binits{V.}},
\bauthor{\bsnm{Guo}, \binits{Y.}}:
\bctitle{Self-supervised detection of contextual synonyms in a multi-class
  setting: Phenotype annotation use case}.
In: \bbtitle{Proceedings of the 2021 Conference on Empirical Methods in Natural
  Language Processing},
pp. \bfpage{8754}--\blpage{8769}.
\bpublisher{Association for Computational Linguistics},
\blocation{Online and Punta Cana, Dominican Republic}
(\byear{2021}).
\burl{https://aclanthology.org/2021.emnlp-main.690}
\end{bchapter}
\endbibitem

\bibitem{johnson2016mimic}
\begin{barticle}
\bauthor{\bsnm{Johnson}, \binits{A.E.}},
\bauthor{\bsnm{Pollard}, \binits{T.J.}},
\bauthor{\bsnm{Shen}, \binits{L.}},
\bauthor{\bsnm{Li-Wei}, \binits{H.L.}},
\bauthor{\bsnm{Feng}, \binits{M.}},
\bauthor{\bsnm{Ghassemi}, \binits{M.}},
\bauthor{\bsnm{Moody}, \binits{B.}},
\bauthor{\bsnm{Szolovits}, \binits{P.}},
\bauthor{\bsnm{Celi}, \binits{L.A.}},
\bauthor{\bsnm{Mark}, \binits{R.G.}}:
\batitle{Mimic-iii, a freely accessible critical care database}.
\bjtitle{Scientific data}
\bvolume{3}(\bissue{1}),
\bfpage{1}--\blpage{9}
(\byear{2016})
\end{barticle}
\endbibitem

\bibitem{harutyunyan2019multitask}
\begin{barticle}
\bauthor{\bsnm{Harutyunyan}, \binits{H.}},
\bauthor{\bsnm{Khachatrian}, \binits{H.}},
\bauthor{\bsnm{Kale}, \binits{D.C.}},
\bauthor{\bsnm{Ver~Steeg}, \binits{G.}},
\bauthor{\bsnm{Galstyan}, \binits{A.}}:
\batitle{Multitask learning and benchmarking with clinical time series data}.
\bjtitle{Scientific data}
\bvolume{6}(\bissue{1}),
\bfpage{1}--\blpage{18}
(\byear{2019})
\end{barticle}
\endbibitem

\bibitem{aerts2006gene}
\begin{barticle}
\bauthor{\bsnm{Aerts}, \binits{S.}},
\bauthor{\bsnm{Lambrechts}, \binits{D.}},
\bauthor{\bsnm{Maity}, \binits{S.}},
\bauthor{\bsnm{Van~Loo}, \binits{P.}},
\bauthor{\bsnm{Coessens}, \binits{B.}},
\bauthor{\bsnm{De~Smet}, \binits{F.}},
\bauthor{\bsnm{Tranchevent}, \binits{L.-C.}},
\bauthor{\bsnm{De~Moor}, \binits{B.}},
\bauthor{\bsnm{Marynen}, \binits{P.}},
\bauthor{\bsnm{Hassan}, \binits{B.}}, \betal:
\batitle{Gene prioritization through genomic data fusion}.
\bjtitle{Nature biotechnology}
\bvolume{24}(\bissue{5}),
\bfpage{537}
(\byear{2006})
\end{barticle}
\endbibitem

\bibitem{son2018deep}
\begin{barticle}
\bauthor{\bsnm{Son}, \binits{J.H.}},
\bauthor{\bsnm{Xie}, \binits{G.}},
\bauthor{\bsnm{Yuan}, \binits{C.}},
\bauthor{\bsnm{Ena}, \binits{L.}},
\bauthor{\bsnm{Li}, \binits{Z.}},
\bauthor{\bsnm{Goldstein}, \binits{A.}},
\bauthor{\bsnm{Huang}, \binits{L.}},
\bauthor{\bsnm{Wang}, \binits{L.}},
\bauthor{\bsnm{Shen}, \binits{F.}},
\bauthor{\bsnm{Liu}, \binits{H.}}, \betal:
\batitle{Deep phenotyping on electronic health records facilitates genetic
  diagnosis by clinical exomes}.
\bjtitle{The American Journal of Human Genetics}
\bvolume{103}(\bissue{1}),
\bfpage{58}--\blpage{73}
(\byear{2018})
\end{barticle}
\endbibitem

\bibitem{Liu2019}
\begin{barticle}
\bauthor{\bsnm{Liu}, \binits{C.}},
\bauthor{\bsnm{Ta}, \binits{C.N.}},
\bauthor{\bsnm{Rogers}, \binits{J.R.}},
\bauthor{\bsnm{Li}, \binits{Z.}},
\bauthor{\bsnm{Lee}, \binits{J.}},
\bauthor{\bsnm{Butler}, \binits{A.M.}},
\bauthor{\bsnm{Shang}, \binits{N.}},
\bauthor{\bsnm{Kury}, \binits{F.S.P.}},
\bauthor{\bsnm{Wang}, \binits{L.}},
\bauthor{\bsnm{Shen}, \binits{F.}},
\bauthor{\bsnm{Liu}, \binits{H.}},
\bauthor{\bsnm{Ena}, \binits{L.}},
\bauthor{\bsnm{Friedman}, \binits{C.}},
\bauthor{\bsnm{Weng}, \binits{C.}}:
\batitle{{Ensembles of natural language processing systems for portable
  phenotyping solutions}}.
\bjtitle{Journal of Biomedical Informatics}
\bvolume{100},
\bfpage{103318}
(\byear{2019}).
doi:\doiurl{10.1016/j.jbi.2019.103318}
\end{barticle}
\endbibitem

\bibitem{Xu2020}
\begin{barticle}
\bauthor{\bsnm{Xu}, \binits{Z.}},
\bauthor{\bsnm{Chou}, \binits{J.}},
\bauthor{\bsnm{Zhang}, \binits{X.S.}},
\bauthor{\bsnm{Luo}, \binits{Y.}},
\bauthor{\bsnm{Isakova}, \binits{T.}},
\bauthor{\bsnm{Adekkanattu}, \binits{P.}},
\bauthor{\bsnm{Ancker}, \binits{J.S.}},
\bauthor{\bsnm{Jiang}, \binits{G.}},
\bauthor{\bsnm{Kiefer}, \binits{R.C.}},
\bauthor{\bsnm{Pacheco}, \binits{J.A.}},
\bauthor{\bsnm{Rasmussen}, \binits{L.V.}},
\bauthor{\bsnm{Pathak}, \binits{J.}},
\bauthor{\bsnm{Wang}, \binits{F.}}:
\batitle{{Identifying sub-phenotypes of acute kidney injury using structured
  and unstructured electronic health record data with memory networks.}}
\bjtitle{Journal of biomedical informatics}
\bvolume{102},
\bfpage{103361}
(\byear{2020}).
doi:\doiurl{10.1016/j.jbi.2019.103361}
\end{barticle}
\endbibitem

\bibitem{zhang2021clinical}
\begin{botherref}
\oauthor{\bsnm{Zhang}, \binits{J.}},
\oauthor{\bsnm{Bolanos}, \binits{L.}},
\oauthor{\bsnm{Tanwar}, \binits{A.}},
\oauthor{\bsnm{Sokol}, \binits{A.}},
\oauthor{\bsnm{Ive}, \binits{J.}},
\oauthor{\bsnm{Gupta}, \binits{V.}},
\oauthor{\bsnm{Guo}, \binits{Y.}}:
{Clinical Utility of the Automatic Phenotype Annotation in Unstructured
  Clinical Notes: ICU Use Cases}.
arXiv preprint arXiv:2107.11665
(2021)
\end{botherref}
\endbibitem

\bibitem{DBLP:journals/nar/0001GMCLVDBBBCC21}
\begin{barticle}
\bauthor{\bsnm{K{\"{o}}hler}, \binits{S.}},
\bauthor{\bsnm{Gargano}, \binits{M.A.}},
\bauthor{\bsnm{Matentzoglu}, \binits{N.}},
\bauthor{\bsnm{Carmody}, \binits{L.}},
\bauthor{\bsnm{Lewis{-}Smith}, \binits{D.}},
\bauthor{\bsnm{Vasilevsky}, \binits{N.A.}},
\bauthor{\bsnm{Danis}, \binits{D.}},
\bauthor{\bsnm{Balagura}, \binits{G.}},
\bauthor{\bsnm{Baynam}, \binits{G.}},
\bauthor{\bsnm{Brower}, \binits{A.M.}},
\bauthor{\bsnm{Callahan}, \binits{T.J.}},
\bauthor{\bsnm{Chute}, \binits{C.G.}},
\bauthor{\bsnm{Est}, \binits{J.L.}},
\bauthor{\bsnm{Galer}, \binits{P.D.}},
\bauthor{\bsnm{Ganesan}, \binits{S.}},
\bauthor{\bsnm{Griese}, \binits{M.}},
\bauthor{\bsnm{Haimel}, \binits{M.}},
\bauthor{\bsnm{Pazmandi}, \binits{J.}},
\bauthor{\bsnm{Hanauer}, \binits{M.}},
\bauthor{\bsnm{Harris}, \binits{N.L.}},
\bauthor{\bsnm{Hartnett}, \binits{M.}},
\bauthor{\bsnm{Hastreiter}, \binits{M.}},
\bauthor{\bsnm{Hauck}, \binits{F.}},
\bauthor{\bsnm{He}, \binits{Y.}},
\bauthor{\bsnm{Jeske}, \binits{T.}},
\bauthor{\bsnm{Kearney}, \binits{H.}},
\bauthor{\bsnm{Kindle}, \binits{G.}},
\bauthor{\bsnm{Klein}, \binits{C.}},
\bauthor{\bsnm{Knoflach}, \binits{K.}},
\bauthor{\bsnm{Krause}, \binits{R.}},
\bauthor{\bsnm{Lagorce}, \binits{D.}},
\bauthor{\bsnm{McMurry}, \binits{J.A.}},
\bauthor{\bsnm{Miller}, \binits{J.A.}},
\bauthor{\bsnm{Munoz{-}Torres}, \binits{M.C.}},
\bauthor{\bsnm{Peters}, \binits{R.L.}},
\bauthor{\bsnm{Rapp}, \binits{C.K.}},
\bauthor{\bsnm{Rath}, \binits{A.}},
\bauthor{\bsnm{Rind}, \binits{S.A.}},
\bauthor{\bsnm{Rosenberg}, \binits{A.Z.}},
\bauthor{\bsnm{Segal}, \binits{M.M.}},
\bauthor{\bsnm{Seidel}, \binits{M.G.}},
\bauthor{\bsnm{Smedley}, \binits{D.}},
\bauthor{\bsnm{Talmy}, \binits{T.}},
\bauthor{\bsnm{Thomas}, \binits{Y.}},
\bauthor{\bsnm{Wiafe}, \binits{S.A.}},
\bauthor{\bsnm{Xian}, \binits{J.}},
\bauthor{\bsnm{Y{\"{u}}ksel}, \binits{Z.}},
\bauthor{\bsnm{Helbig}, \binits{I.}},
\bauthor{\bsnm{Mungall}, \binits{C.J.}},
\bauthor{\bsnm{Haendel}, \binits{M.A.}},
\bauthor{\bsnm{Robinson}, \binits{P.N.}}:
\batitle{{The Human Phenotype Ontology in 2021}}.
\bjtitle{Nucleic Acids Res.}
\bvolume{49}(\bissue{Database-Issue}),
\bfpage{1207}--\blpage{1217}
(\byear{2021}).
doi:\doiurl{10.1093/nar/gkaa1043}
\end{barticle}
\endbibitem

\bibitem{vasvani:nips:2018}
\begin{bchapter}
\bauthor{\bsnm{Vaswani}, \binits{A.}},
\bauthor{\bsnm{Shazeer}, \binits{N.}},
\bauthor{\bsnm{Parmar}, \binits{N.}},
\bauthor{\bsnm{Uszkoreit}, \binits{J.}},
\bauthor{\bsnm{Jones}, \binits{L.}},
\bauthor{\bsnm{Gomez}, \binits{A.N.}},
\bauthor{\bsnm{Kaiser}, \binits{L.}},
\bauthor{\bsnm{Polosukhin}, \binits{I.}}:
\bctitle{Attention is all you need}.
In: \bbtitle{Advances in Neural Information Processing Systems 30},
pp. \bfpage{5998}--\blpage{6008}
(\byear{2017}).
\burl{http://papers.nips.cc/paper/7181-attention-is-all-you-need.pdf}
\end{bchapter}
\endbibitem

\bibitem{demner2017metamaplite}
\begin{barticle}
\bauthor{\bsnm{Demner-Fushman}, \binits{D.}},
\bauthor{\bsnm{Rogers}, \binits{W.J.}},
\bauthor{\bsnm{Aronson}, \binits{A.R.}}:
\batitle{Metamap lite: an evaluation of a new java implementation of metamap}.
\bjtitle{Journal of the American Medical Informatics Association}
\bvolume{24}(\bissue{4}),
\bfpage{841}--\blpage{844}
(\byear{2017})
\end{barticle}
\endbibitem

\bibitem{arbabi2019ncr}
\begin{barticle}
\bauthor{\bsnm{Arbabi}, \binits{A.}},
\bauthor{\bsnm{Adams}, \binits{D.R.}},
\bauthor{\bsnm{Fidler}, \binits{S.}},
\bauthor{\bsnm{Brudno}, \binits{M.}}:
\batitle{{Identifying clinical terms in medical text using Ontology-Guided
  machine learning}}.
\bjtitle{JMIR medical informatics}
\bvolume{7}(\bissue{2}),
\bfpage{12596}
(\byear{2019})
\end{barticle}
\endbibitem

\bibitem{kraljevic2019medcat}
\begin{botherref}
\oauthor{\bsnm{Kraljevic}, \binits{Z.}},
\oauthor{\bsnm{Bean}, \binits{D.}},
\oauthor{\bsnm{Mascio}, \binits{A.}},
\oauthor{\bsnm{Roguski}, \binits{L.}},
\oauthor{\bsnm{Folarin}, \binits{A.}},
\oauthor{\bsnm{Roberts}, \binits{A.}},
\oauthor{\bsnm{Bendayan}, \binits{R.}},
\oauthor{\bsnm{Dobson}, \binits{R.}}:
MedCAT -- Medical Concept Annotation Tool
(2019).
\arxivurl{1912.10166}
\end{botherref}
\endbibitem

\bibitem{devlin-etal-2019-bert}
\begin{bchapter}
\bauthor{\bsnm{Devlin}, \binits{J.}},
\bauthor{\bsnm{Chang}, \binits{M.-W.}},
\bauthor{\bsnm{Lee}, \binits{K.}},
\bauthor{\bsnm{Toutanova}, \binits{K.}}:
\bctitle{{BERT}: Pre-training of deep bidirectional transformers for language
  understanding}.
In: \bbtitle{Proceedings of the 2019 Conference of the North {A}merican Chapter
  of the Association for Computational Linguistics: Human Language
  Technologies, Volume 1 (Long and Short Papers)},
pp. \bfpage{4171}--\blpage{4186}.
\bpublisher{Association for Computational Linguistics},
\blocation{Minneapolis, Minnesota}
(\byear{2019}).
doi:\doiurl{10.18653/v1/N19-1423}.
\burl{https://aclanthology.org/N19-1423}
\end{bchapter}
\endbibitem

\bibitem{biobert}
\begin{barticle}
\bauthor{\bsnm{Lee}, \binits{J.}},
\bauthor{\bsnm{Yoon}, \binits{W.}},
\bauthor{\bsnm{Kim}, \binits{S.}},
\bauthor{\bsnm{Kim}, \binits{D.}},
\bauthor{\bsnm{Kim}, \binits{S.}},
\bauthor{\bsnm{So}, \binits{C.H.}},
\bauthor{\bsnm{Kang}, \binits{J.}}:
\batitle{{BioBERT: a pre-trained biomedical language representation model for
  biomedical text mining}}.
\bjtitle{Bioinformatics}
(\byear{2019}).
doi:\doiurl{10.1093/bioinformatics/btz682}
\end{barticle}
\endbibitem

\bibitem{alsentzer-etal-2019-publicly}
\begin{bchapter}
\bauthor{\bsnm{Alsentzer}, \binits{E.}},
\bauthor{\bsnm{Murphy}, \binits{J.}},
\bauthor{\bsnm{Boag}, \binits{W.}},
\bauthor{\bsnm{Weng}, \binits{W.-H.}},
\bauthor{\bsnm{Jindi}, \binits{D.}},
\bauthor{\bsnm{Naumann}, \binits{T.}},
\bauthor{\bsnm{McDermott}, \binits{M.}}:
\bctitle{Publicly available clinical {BERT} embeddings}.
In: \bbtitle{Proceedings of the 2nd Clinical Natural Language Processing
  Workshop},
pp. \bfpage{72}--\blpage{78}.
\bpublisher{Association for Computational Linguistics},
\blocation{Minneapolis, Minnesota, USA}
(\byear{2019}).
doi:\doiurl{10.18653/v1/W19-1909}.
\burl{https://aclanthology.org/W19-1909}
\end{bchapter}
\endbibitem

\bibitem{beltagy-etal-2019-scibert}
\begin{bchapter}
\bauthor{\bsnm{Beltagy}, \binits{I.}},
\bauthor{\bsnm{Lo}, \binits{K.}},
\bauthor{\bsnm{Cohan}, \binits{A.}}:
\bctitle{{S}ci{BERT}: A pretrained language model for scientific text}.
In: \bbtitle{Proceedings of the 2019 Conference on Empirical Methods in Natural
  Language Processing and the 9th International Joint Conference on Natural
  Language Processing (EMNLP-IJCNLP)},
pp. \bfpage{3615}--\blpage{3620}.
\bpublisher{Association for Computational Linguistics},
\blocation{Hong Kong, China}
(\byear{2019}).
doi:\doiurl{10.18653/v1/D19-1371}.
\burl{https://aclanthology.org/D19-1371}
\end{bchapter}
\endbibitem

\bibitem{he2021automl}
\begin{barticle}
\bauthor{\bsnm{He}, \binits{X.}},
\bauthor{\bsnm{Zhao}, \binits{K.}},
\bauthor{\bsnm{Chu}, \binits{X.}}:
\batitle{Automl: A survey of the state-of-the-art}.
\bjtitle{Knowledge-Based Systems}
\bvolume{212},
\bfpage{106622}
(\byear{2021})
\end{barticle}
\endbibitem

\bibitem{falkner-icml-18}
\begin{bchapter}
\bauthor{\bsnm{Falkner}, \binits{S.}},
\bauthor{\bsnm{Klein}, \binits{A.}},
\bauthor{\bsnm{Hutter}, \binits{F.}}:
\bctitle{{BOHB}: Robust and efficient hyperparameter optimization at scale}.
In: \bbtitle{Proceedings of the 35th International Conference on Machine
  Learning},
pp. \bfpage{1436}--\blpage{1445}
(\byear{2018})
\end{bchapter}
\endbibitem

\bibitem{scikit-learn}
\begin{barticle}
\bauthor{\bsnm{Pedregosa}, \binits{F.}},
\bauthor{\bsnm{Varoquaux}, \binits{G.}},
\bauthor{\bsnm{Gramfort}, \binits{A.}},
\bauthor{\bsnm{Michel}, \binits{V.}},
\bauthor{\bsnm{Thirion}, \binits{B.}},
\bauthor{\bsnm{Grisel}, \binits{O.}},
\bauthor{\bsnm{Blondel}, \binits{M.}},
\bauthor{\bsnm{Prettenhofer}, \binits{P.}},
\bauthor{\bsnm{Weiss}, \binits{R.}},
\bauthor{\bsnm{Dubourg}, \binits{V.}},
\bauthor{\bsnm{Vanderplas}, \binits{J.}},
\bauthor{\bsnm{Passos}, \binits{A.}},
\bauthor{\bsnm{Cournapeau}, \binits{D.}},
\bauthor{\bsnm{Brucher}, \binits{M.}},
\bauthor{\bsnm{Perrot}, \binits{M.}},
\bauthor{\bsnm{Duchesnay}, \binits{E.}}:
\batitle{Scikit-learn: Machine learning in {P}ython}.
\bjtitle{Journal of Machine Learning Research}
\bvolume{12},
\bfpage{2825}--\blpage{2830}
(\byear{2011})
\end{barticle}
\endbibitem

\bibitem{SHAP}
\begin{botherref}
\oauthor{\bsnm{Lundberg}, \binits{S.M.}},
\oauthor{\bsnm{Lee}, \binits{S.-I.}}:
A unified approach to interpreting model predictions.
Advances in neural information processing systems
\textbf{30}
(2017)
\end{botherref}
\endbibitem

\bibitem{Breiman2001}
\begin{barticle}
\bauthor{\bsnm{Breiman}, \binits{L.}}:
\batitle{Random forests}.
\bjtitle{Machine Learning}
\bvolume{45},
\bfpage{5}--\blpage{32}
(\byear{2001}).
doi:\doiurl{10.1023/A:1010933404324}
\end{barticle}
\endbibitem

\end{thebibliography}

\newcommand{\BMCxmlcomment}[1]{}

\BMCxmlcomment{

<refgrp>

<bibl id="B1">
  <title><p>Activity in the NHS</p></title>
  <pubdate>2020</pubdate>
  <url>https://www.kingsfund.org.uk/projects/nhs-in-a-nutshell/NHS-activity</url>
</bibl>

<bibl id="B2">
  <title><p>Big data in healthcare: management, analysis and future
  prospects</p></title>
  <aug>
    <au><snm>Dash</snm><fnm>S</fnm></au>
    <au><snm>Shakyawar</snm><fnm>SK</fnm></au>
    <au><snm>Sharma</snm><fnm>M</fnm></au>
    <au><snm>Kaushik</snm><fnm>S</fnm></au>
  </aug>
  <source>Journal of Big Data</source>
  <publisher>Springer Science and Business Media {LLC}</publisher>
  <pubdate>2019</pubdate>
  <volume>6</volume>
  <issue>1</issue>
  <url>https://doi.org/10.1186/s40537-019-0217-0</url>
</bibl>

<bibl id="B3">
  <title><p>Applications of machine learning in drug discovery and
  development</p></title>
  <aug>
    <au><snm>Vamathevan</snm><fnm>J</fnm></au>
    <au><snm>Clark</snm><fnm>D</fnm></au>
    <au><snm>Czodrowski</snm><fnm>P</fnm></au>
    <au><snm>Dunham</snm><fnm>I</fnm></au>
    <au><snm>Ferran</snm><fnm>E</fnm></au>
    <au><snm>Lee</snm><fnm>G</fnm></au>
    <au><snm>Li</snm><fnm>B</fnm></au>
    <au><snm>Madabhushi</snm><fnm>A</fnm></au>
    <au><snm>Shah</snm><fnm>P</fnm></au>
    <au><snm>Spitzer</snm><fnm>M</fnm></au>
    <au><snm>Zhao</snm><fnm>S</fnm></au>
  </aug>
  <source>Nature Reviews Drug Discovery</source>
  <publisher>Springer Science and Business Media {LLC}</publisher>
  <pubdate>2019</pubdate>
  <volume>18</volume>
  <issue>6</issue>
  <fpage>463</fpage>
  <lpage>-477</lpage>
  <url>https://doi.org/10.1038/s41573-019-0024-5</url>
</bibl>

<bibl id="B4">
  <title><p>Development of machine learning model for diagnostic disease
  prediction based on laboratory tests</p></title>
  <aug>
    <au><snm>Park</snm><fnm>DJ</fnm></au>
    <au><snm>Park</snm><fnm>MW</fnm></au>
    <au><snm>Lee</snm><fnm>H</fnm></au>
    <au><snm>Kim</snm><fnm>YJ</fnm></au>
    <au><snm>Kim</snm><fnm>Y</fnm></au>
    <au><snm>Park</snm><fnm>YH</fnm></au>
  </aug>
  <source>Scientific Reports</source>
  <publisher>Springer Science and Business Media {LLC}</publisher>
  <pubdate>2021</pubdate>
  <volume>11</volume>
  <issue>1</issue>
  <url>https://doi.org/10.1038/s41598-021-87171-5</url>
</bibl>

<bibl id="B5">
  <title><p>Untangling the complexity of multimorbidity with machine
  learning</p></title>
  <aug>
    <au><snm>Hassaine</snm><fnm>A</fnm></au>
    <au><snm>Salimi Khorshidi</snm><fnm>G</fnm></au>
    <au><snm>Canoy</snm><fnm>D</fnm></au>
    <au><snm>Rahimi</snm><fnm>K</fnm></au>
  </aug>
  <source>Mechanisms of Ageing and Development</source>
  <publisher>Elsevier {BV}</publisher>
  <pubdate>2020</pubdate>
  <volume>190</volume>
  <fpage>111325</fpage>
  <url>https://doi.org/10.1016/j.mad.2020.111325</url>
</bibl>

<bibl id="B6">
  <title><p>Three- and four-digit {ICD}-10 is not a reliable classification
  system in primary care</p></title>
  <aug>
    <au><snm>Wockenfuss</snm><fnm>R</fnm></au>
    <au><snm>Frese</snm><fnm>T</fnm></au>
    <au><snm>Herrmann</snm><fnm>K</fnm></au>
    <au><snm>Claussnitzer</snm><fnm>M</fnm></au>
    <au><snm>Sandholzer</snm><fnm>H</fnm></au>
  </aug>
  <source>Scandinavian Journal of Primary Health Care</source>
  <publisher>Informa {UK} Limited</publisher>
  <pubdate>2009</pubdate>
  <volume>27</volume>
  <issue>3</issue>
  <fpage>131</fpage>
  <lpage>-136</lpage>
  <url>https://doi.org/10.1080/02813430903072215</url>
</bibl>

<bibl id="B7">
  <title><p>{Opportunities and challenges for quality and safety applications
  in ICD-11: an international survey of users of coded health data}</p></title>
  <aug>
    <au><snm>Southern</snm><fnm>DA</fnm></au>
    <au><snm>Hall</snm><fnm>M</fnm></au>
    <au><snm>White</snm><fnm>DE</fnm></au>
    <au><snm>Romano</snm><fnm>PS</fnm></au>
    <au><snm>Sundararajan</snm><fnm>V</fnm></au>
    <au><snm>Droesler</snm><fnm>SE</fnm></au>
    <au><snm>Pincus</snm><fnm>HA</fnm></au>
    <au><snm>Ghali</snm><fnm>WA</fnm></au>
  </aug>
  <source>International Journal for Quality in Health Care</source>
  <pubdate>2015</pubdate>
  <volume>28</volume>
  <issue>1</issue>
  <fpage>129</fpage>
  <lpage>135</lpage>
  <url>https://doi.org/10.1093/intqhc/mzv096</url>
</bibl>

<bibl id="B8">
  <title><p>The accuracy of hospital ICD-9-CM codes for determining Sickle Cell
  Disease genotype</p></title>
  <aug>
    <au><snm>Snyder</snm><fnm>AB</fnm></au>
    <au><snm>Lane</snm><fnm>PA</fnm></au>
    <au><snm>Zhou</snm><fnm>M</fnm></au>
    <au><snm>Paulukonis</snm><fnm>ST</fnm></au>
    <au><snm>Hulihan</snm><fnm>MM</fnm></au>
  </aug>
  <source>Journal of rare diseases research {\&} treatment</source>
  <edition>20170728</edition>
  <pubdate>2017</pubdate>
  <volume>2</volume>
  <issue>4</issue>
  <fpage>39</fpage>
  <lpage>45</lpage>
  <url>https://pubmed.ncbi.nlm.nih.gov/29202133</url>
  <note>29202133[pmid]</note>
</bibl>

<bibl id="B9">
  <title><p>Accuracy and Completeness of Clinical Coding Using ICD-10 for
  Ambulatory Visits</p></title>
  <aug>
    <au><snm>Horsky</snm><fnm>J</fnm></au>
    <au><snm>Drucker</snm><fnm>EA</fnm></au>
    <au><snm>Ramelson</snm><fnm>HZ</fnm></au>
  </aug>
  <source>AMIA ... Annual Symposium proceedings. AMIA Symposium</source>
  <publisher>American Medical Informatics Association</publisher>
  <pubdate>2018</pubdate>
  <volume>2017</volume>
  <fpage>912</fpage>
  <lpage>920</lpage>
  <url>https://pubmed.ncbi.nlm.nih.gov/29854158</url>
  <note>29854158[pmid]</note>
</bibl>

<bibl id="B10">
  <title><p>Limitations of pulmonary embolism {ICD}-10 codes in emergency
  department administrative data: let the buyer beware</p></title>
  <aug>
    <au><snm>Burles</snm><fnm>K</fnm></au>
    <au><snm>Innes</snm><fnm>G</fnm></au>
    <au><snm>Senior</snm><fnm>K</fnm></au>
    <au><snm>Lang</snm><fnm>E</fnm></au>
    <au><snm>McRae</snm><fnm>A</fnm></au>
  </aug>
  <source>{BMC} Medical Research Methodology</source>
  <publisher>Springer Science and Business Media {LLC}</publisher>
  <pubdate>2017</pubdate>
  <volume>17</volume>
  <issue>1</issue>
  <url>https://doi.org/10.1186/s12874-017-0361-1</url>
</bibl>

<bibl id="B11">
  <title><p>Identifying sub-phenotypes of acute kidney injury using structured
  and unstructured electronic health record data with memory
  networks</p></title>
  <aug>
    <au><snm>Xu</snm><fnm>Z</fnm></au>
    <au><snm>Chou</snm><fnm>J</fnm></au>
    <au><snm>Zhang</snm><fnm>XS</fnm></au>
    <au><snm>Luo</snm><fnm>Y</fnm></au>
    <au><snm>Isakova</snm><fnm>T</fnm></au>
    <au><snm>Adekkanattu</snm><fnm>P</fnm></au>
    <au><snm>Ancker</snm><fnm>JS</fnm></au>
    <au><snm>Jiang</snm><fnm>G</fnm></au>
    <au><snm>Kiefer</snm><fnm>RC</fnm></au>
    <au><snm>Pacheco</snm><fnm>JA</fnm></au>
    <au><cnm>others</cnm></au>
  </aug>
  <source>Journal of biomedical informatics</source>
  <publisher>Elsevier</publisher>
  <pubdate>2020</pubdate>
  <volume>102</volume>
  <fpage>103361</fpage>
</bibl>

<bibl id="B12">
  <title><p>Evaluating eligibility criteria of oncology trials using real-world
  data and AI</p></title>
  <aug>
    <au><snm>Liu</snm><fnm>R</fnm></au>
    <au><snm>Rizzo</snm><fnm>S</fnm></au>
    <au><snm>Whipple</snm><fnm>S</fnm></au>
    <au><snm>Pal</snm><fnm>N</fnm></au>
    <au><snm>Pineda</snm><fnm>AL</fnm></au>
    <au><snm>Lu</snm><fnm>M</fnm></au>
    <au><snm>Arnieri</snm><fnm>B</fnm></au>
    <au><snm>Lu</snm><fnm>Y</fnm></au>
    <au><snm>Capra</snm><fnm>W</fnm></au>
    <au><snm>Copping</snm><fnm>R</fnm></au>
    <au><cnm>others</cnm></au>
  </aug>
  <source>Nature</source>
  <publisher>Nature Publishing Group</publisher>
  <pubdate>2021</pubdate>
  <volume>592</volume>
  <issue>7855</issue>
  <fpage>629</fpage>
  <lpage>-633</lpage>
</bibl>

<bibl id="B13">
  <title><p>Prediction of fatty liver disease using machine learning
  algorithms</p></title>
  <aug>
    <au><snm>Wu</snm><fnm>CC</fnm></au>
    <au><snm>Yeh</snm><fnm>WC</fnm></au>
    <au><snm>Hsu</snm><fnm>WD</fnm></au>
    <au><snm>Islam</snm><fnm>MM</fnm></au>
    <au><snm>Nguyen</snm><fnm>PAA</fnm></au>
    <au><snm>Poly</snm><fnm>TN</fnm></au>
    <au><snm>Wang</snm><fnm>YC</fnm></au>
    <au><snm>Yang</snm><fnm>HC</fnm></au>
    <au><snm>Li</snm><fnm>YCJ</fnm></au>
  </aug>
  <source>Computer methods and programs in biomedicine</source>
  <publisher>Elsevier</publisher>
  <pubdate>2019</pubdate>
  <volume>170</volume>
  <fpage>23</fpage>
  <lpage>-29</lpage>
</bibl>

<bibl id="B14">
  <title><p>Detecting undiagnosed atrial fibrillation in UK primary care:
  validation of a machine learning prediction algorithm in a retrospective
  cohort study</p></title>
  <aug>
    <au><snm>Sekelj</snm><fnm>S</fnm></au>
    <au><snm>Sandler</snm><fnm>B</fnm></au>
    <au><snm>Johnston</snm><fnm>E</fnm></au>
    <au><snm>Pollock</snm><fnm>KG</fnm></au>
    <au><snm>Hill</snm><fnm>NR</fnm></au>
    <au><snm>Gordon</snm><fnm>J</fnm></au>
    <au><snm>Tsang</snm><fnm>C</fnm></au>
    <au><snm>Khan</snm><fnm>S</fnm></au>
    <au><snm>Ng</snm><fnm>FS</fnm></au>
    <au><snm>Farooqui</snm><fnm>U</fnm></au>
  </aug>
  <source>European journal of preventive cardiology</source>
  <publisher>Oxford University Press</publisher>
  <pubdate>2021</pubdate>
  <volume>28</volume>
  <issue>6</issue>
  <fpage>598</fpage>
  <lpage>-605</lpage>
</bibl>

<bibl id="B15">
  <title><p>Machine Learning for Prediction and Risk Stratification of Lupus
  Nephritis Renal Flare</p></title>
  <aug>
    <au><snm>Chen</snm><fnm>Y</fnm></au>
    <au><snm>Huang</snm><fnm>S</fnm></au>
    <au><snm>Chen</snm><fnm>T</fnm></au>
    <au><snm>Liang</snm><fnm>D</fnm></au>
    <au><snm>Yang</snm><fnm>J</fnm></au>
    <au><snm>Zeng</snm><fnm>C</fnm></au>
    <au><snm>Li</snm><fnm>X</fnm></au>
    <au><snm>Xie</snm><fnm>G</fnm></au>
    <au><snm>Liu</snm><fnm>Z</fnm></au>
  </aug>
  <source>American Journal of Nephrology</source>
  <publisher>S. Karger {AG}</publisher>
  <pubdate>2021</pubdate>
  <volume>52</volume>
  <issue>2</issue>
  <fpage>152</fpage>
  <lpage>-160</lpage>
  <url>https://doi.org/10.1159/000513566</url>
</bibl>

<bibl id="B16">
  <title><p>Machine learning approaches to predict lupus disease activity from
  gene expression data</p></title>
  <aug>
    <au><snm>Kegerreis</snm><fnm>B</fnm></au>
    <au><snm>Catalina</snm><fnm>MD</fnm></au>
    <au><snm>Bachali</snm><fnm>P</fnm></au>
    <au><snm>Geraci</snm><fnm>NS</fnm></au>
    <au><snm>Labonte</snm><fnm>AC</fnm></au>
    <au><snm>Zeng</snm><fnm>C</fnm></au>
    <au><snm>Stearrett</snm><fnm>N</fnm></au>
    <au><snm>Crandall</snm><fnm>KA</fnm></au>
    <au><snm>Lipsky</snm><fnm>PE</fnm></au>
    <au><snm>Grammer</snm><fnm>AC</fnm></au>
  </aug>
  <source>Scientific Reports</source>
  <publisher>Springer Science and Business Media {LLC}</publisher>
  <pubdate>2019</pubdate>
  <volume>9</volume>
  <issue>1</issue>
  <url>https://doi.org/10.1038/s41598-019-45989-0</url>
</bibl>

<bibl id="B17">
  <title><p>Combining billing codes, clinical notes, and medications from
  electronic health records provides superior phenotyping
  performance</p></title>
  <aug>
    <au><snm>Wei</snm><fnm>WQ</fnm></au>
    <au><snm>Teixeira</snm><fnm>PL</fnm></au>
    <au><snm>Mo</snm><fnm>H</fnm></au>
    <au><snm>Cronin</snm><fnm>RM</fnm></au>
    <au><snm>Warner</snm><fnm>JL</fnm></au>
    <au><snm>Denny</snm><fnm>JC</fnm></au>
  </aug>
  <source>Journal of the American Medical Informatics Association</source>
  <publisher>Oxford University Press ({OUP})</publisher>
  <pubdate>2015</pubdate>
  <volume>23</volume>
  <issue>e1</issue>
  <fpage>e20</fpage>
  <lpage>-e27</lpage>
  <url>https://doi.org/10.1093/jamia/ocv130</url>
</bibl>

<bibl id="B18">
  <title><p>Managing Unstructured Big Data in Healthcare System</p></title>
  <aug>
    <au><snm>Kong</snm><fnm>HJ</fnm></au>
  </aug>
  <source>Healthcare Informatics Research</source>
  <publisher>The Korean Society of Medical Informatics</publisher>
  <pubdate>2019</pubdate>
  <volume>25</volume>
  <issue>1</issue>
  <fpage>1</fpage>
  <url>https://doi.org/10.4258/hir.2019.25.1.1</url>
</bibl>

<bibl id="B19">
  <title><p>Multimodal machine learning for automated ICD coding</p></title>
  <aug>
    <au><snm>Xu</snm><fnm>K</fnm></au>
    <au><snm>Lam</snm><fnm>M</fnm></au>
    <au><snm>Pang</snm><fnm>J</fnm></au>
    <au><snm>Gao</snm><fnm>X</fnm></au>
    <au><snm>Band</snm><fnm>C</fnm></au>
    <au><snm>Mathur</snm><fnm>P</fnm></au>
    <au><snm>Papay</snm><fnm>F</fnm></au>
    <au><snm>Khanna</snm><fnm>AK</fnm></au>
    <au><snm>Cywinski</snm><fnm>JB</fnm></au>
    <au><snm>Maheshwari</snm><fnm>K</fnm></au>
    <au><cnm>others</cnm></au>
  </aug>
  <source>Machine Learning for Healthcare Conference</source>
  <pubdate>2019</pubdate>
  <fpage>197</fpage>
  <lpage>-215</lpage>
</bibl>

<bibl id="B20">
  <title><p>{BERT}-{XML}: Large Scale Automated {ICD} Coding Using {BERT}
  Pretraining</p></title>
  <aug>
    <au><snm>Zhang</snm><fnm>Z</fnm></au>
    <au><snm>Liu</snm><fnm>J</fnm></au>
    <au><snm>Razavian</snm><fnm>N</fnm></au>
  </aug>
  <source>Proceedings of the 3rd Clinical Natural Language Processing
  Workshop</source>
  <publisher>Online: Association for Computational Linguistics</publisher>
  <pubdate>2020</pubdate>
  <fpage>24</fpage>
  <lpage>-34</lpage>
  <url>https://aclanthology.org/2020.clinicalnlp-1.3</url>
</bibl>

<bibl id="B21">
  <title><p>Deep phenotyping for precision medicine</p></title>
  <aug>
    <au><snm>Robinson</snm><fnm>PN</fnm></au>
  </aug>
  <source>Human mutation</source>
  <publisher>Wiley Online Library</publisher>
  <pubdate>2012</pubdate>
  <volume>33</volume>
  <issue>5</issue>
  <fpage>777</fpage>
  <lpage>-780</lpage>
</bibl>

<bibl id="B22">
  <title><p>Enabling phenotypic big data with PheNorm</p></title>
  <aug>
    <au><snm>Yu</snm><fnm>S</fnm></au>
    <au><snm>Ma</snm><fnm>Y</fnm></au>
    <au><snm>Gronsbell</snm><fnm>J</fnm></au>
    <au><snm>Cai</snm><fnm>T</fnm></au>
    <au><snm>Ananthakrishnan</snm><fnm>AN</fnm></au>
    <au><snm>Gainer</snm><fnm>VS</fnm></au>
    <au><snm>Churchill</snm><fnm>SE</fnm></au>
    <au><snm>Szolovits</snm><fnm>P</fnm></au>
    <au><snm>Murphy</snm><fnm>SN</fnm></au>
    <au><snm>Kohane</snm><fnm>IS</fnm></au>
    <au><cnm>others</cnm></au>
  </aug>
  <source>Journal of the American Medical Informatics Association</source>
  <publisher>Oxford University Press</publisher>
  <pubdate>2018</pubdate>
  <volume>25</volume>
  <issue>1</issue>
  <fpage>54</fpage>
  <lpage>-60</lpage>
</bibl>

<bibl id="B23">
  <title><p>2019 European League Against Rheumatism/American College of
  Rheumatology Classification Criteria for Systemic Lupus
  Erythematosus</p></title>
  <aug>
    <au><snm>Aringer</snm><fnm>M</fnm></au>
    <au><snm>Costenbader</snm><fnm>K</fnm></au>
    <au><snm>Daikh</snm><fnm>D</fnm></au>
    <au><snm>Brinks</snm><fnm>R</fnm></au>
    <au><snm>Mosca</snm><fnm>M</fnm></au>
    <au><snm>Ramsey Goldman</snm><fnm>R</fnm></au>
    <au><snm>Smolen</snm><fnm>JS</fnm></au>
    <au><snm>Wofsy</snm><fnm>D</fnm></au>
    <au><snm>Boumpas</snm><fnm>DT</fnm></au>
    <au><snm>Kamen</snm><fnm>DL</fnm></au>
    <au><snm>Jayne</snm><fnm>D</fnm></au>
    <au><snm>Cervera</snm><fnm>R</fnm></au>
    <au><snm>Costedoat Chalumeau</snm><fnm>N</fnm></au>
    <au><snm>Diamond</snm><fnm>B</fnm></au>
    <au><snm>Gladman</snm><fnm>DD</fnm></au>
    <au><snm>Hahn</snm><fnm>B</fnm></au>
    <au><snm>Hiepe</snm><fnm>F</fnm></au>
    <au><snm>Jacobsen</snm><fnm>S</fnm></au>
    <au><snm>Khanna</snm><fnm>D</fnm></au>
    <au><snm>Lerstr{\o}m</snm><fnm>K</fnm></au>
    <au><snm>Massarotti</snm><fnm>E</fnm></au>
    <au><snm>McCune</snm><fnm>J</fnm></au>
    <au><snm>Ruiz Irastorza</snm><fnm>G</fnm></au>
    <au><snm>Sanchez Guerrero</snm><fnm>J</fnm></au>
    <au><snm>Schneider</snm><fnm>M</fnm></au>
    <au><snm>Urowitz</snm><fnm>M</fnm></au>
    <au><snm>Bertsias</snm><fnm>G</fnm></au>
    <au><snm>Hoyer</snm><fnm>BF</fnm></au>
    <au><snm>Leuchten</snm><fnm>N</fnm></au>
    <au><snm>Tani</snm><fnm>C</fnm></au>
    <au><snm>Tedeschi</snm><fnm>SK</fnm></au>
    <au><snm>Touma</snm><fnm>Z</fnm></au>
    <au><snm>Schmajuk</snm><fnm>G</fnm></au>
    <au><snm>Anic</snm><fnm>B</fnm></au>
    <au><snm>Assan</snm><fnm>F</fnm></au>
    <au><snm>Chan</snm><fnm>TM</fnm></au>
    <au><snm>Clarke</snm><fnm>AE</fnm></au>
    <au><snm>Crow</snm><fnm>MK</fnm></au>
    <au><snm>Czirj{\'a}k</snm><fnm>L</fnm></au>
    <au><snm>Doria</snm><fnm>A</fnm></au>
    <au><snm>Graninger</snm><fnm>W</fnm></au>
    <au><snm>Halda Kiss</snm><fnm>B</fnm></au>
    <au><snm>Hasni</snm><fnm>S</fnm></au>
    <au><snm>Izmirly</snm><fnm>PM</fnm></au>
    <au><snm>Jung</snm><fnm>M</fnm></au>
    <au><snm>Kum{\'a}novics</snm><fnm>G</fnm></au>
    <au><snm>Mariette</snm><fnm>X</fnm></au>
    <au><snm>Padjen</snm><fnm>I</fnm></au>
    <au><snm>Pego Reigosa</snm><fnm>JM</fnm></au>
    <au><snm>Romero Diaz</snm><fnm>J</fnm></au>
    <au><snm>R{\'u}a Figueroa Fern{\'a}ndez</snm><fnm>{\'I}</fnm></au>
    <au><snm>Seror</snm><fnm>R</fnm></au>
    <au><snm>Stummvoll</snm><fnm>GH</fnm></au>
    <au><snm>Tanaka</snm><fnm>Y</fnm></au>
    <au><snm>Tektonidou</snm><fnm>MG</fnm></au>
    <au><snm>Vasconcelos</snm><fnm>C</fnm></au>
    <au><snm>Vital</snm><fnm>EM</fnm></au>
    <au><snm>Wallace</snm><fnm>DJ</fnm></au>
    <au><snm>Yavuz</snm><fnm>S</fnm></au>
    <au><snm>Meroni</snm><fnm>PL</fnm></au>
    <au><snm>Fritzler</snm><fnm>MJ</fnm></au>
    <au><snm>Naden</snm><fnm>R</fnm></au>
    <au><snm>D{\"o}rner</snm><fnm>T</fnm></au>
    <au><snm>Johnson</snm><fnm>SR</fnm></au>
  </aug>
  <source>Arthritis {\&} rheumatology (Hoboken, N.J.)</source>
  <edition>20190806</edition>
  <pubdate>2019</pubdate>
  <volume>71</volume>
  <issue>9</issue>
  <fpage>1400</fpage>
  <lpage>1412</lpage>
  <url>https://pubmed.ncbi.nlm.nih.gov/31385462</url>
  <note>31385462[pmid]</note>
</bibl>

<bibl id="B24">
  <title><p>ClinPhen extracts and prioritizes patient phenotypes directly from
  medical records to expedite genetic disease diagnosis</p></title>
  <aug>
    <au><snm>Deisseroth</snm><fnm>CA</fnm></au>
    <au><snm>Birgmeier</snm><fnm>J</fnm></au>
    <au><snm>Bodle</snm><fnm>EE</fnm></au>
    <au><snm>Kohler</snm><fnm>JN</fnm></au>
    <au><snm>Matalon</snm><fnm>DR</fnm></au>
    <au><snm>Nazarenko</snm><fnm>Y</fnm></au>
    <au><snm>Genetti</snm><fnm>CA</fnm></au>
    <au><snm>Brownstein</snm><fnm>CA</fnm></au>
    <au><snm>Schmitz Abe</snm><fnm>K</fnm></au>
    <au><snm>Schoch</snm><fnm>K</fnm></au>
    <au><cnm>others</cnm></au>
  </aug>
  <source>Genetics in Medicine</source>
  <publisher>Nature Publishing Group</publisher>
  <pubdate>2019</pubdate>
  <volume>21</volume>
  <issue>7</issue>
  <fpage>1585</fpage>
  <lpage>-1593</lpage>
</bibl>

<bibl id="B25">
  <title><p>{NCBO Annotator: Semantic Annotation of Biomedical
  Data}</p></title>
  <aug>
    <au><snm>Jonquet</snm><fnm>C</fnm></au>
    <au><snm>Shah</snm><fnm>N</fnm></au>
    <au><snm>Youn</snm><fnm>C</fnm></au>
    <au><snm>Callendar</snm><fnm>C</fnm></au>
    <au><snm>Storey</snm><fnm>MA</fnm></au>
    <au><snm>Musen</snm><fnm>M</fnm></au>
  </aug>
  <source>International Semantic Web Conference, Poster and Demo
  session</source>
  <pubdate>2009</pubdate>
  <volume>110</volume>
</bibl>

<bibl id="B26">
  <title><p>{Mayo clinical Text Analysis and Knowledge Extraction System
  (cTAKES): architecture, component evaluation and applications}</p></title>
  <aug>
    <au><snm>Savova</snm><fnm>GK</fnm></au>
    <au><snm>Masanz</snm><fnm>JJ</fnm></au>
    <au><snm>Ogren</snm><fnm>PV</fnm></au>
    <au><snm>Zheng</snm><fnm>J</fnm></au>
    <au><snm>Sohn</snm><fnm>S</fnm></au>
    <au><snm>Kipper Schuler</snm><fnm>KC</fnm></au>
    <au><snm>Chute</snm><fnm>CG</fnm></au>
  </aug>
  <source>Journal of the American Medical Informatics Association :
  JAMIA</source>
  <publisher>BMA House, Tavistock Square, London, WC1H 9JR: BMJ
  Group</publisher>
  <pubdate>2010</pubdate>
  <volume>17</volume>
  <issue>5</issue>
  <fpage>507</fpage>
  <lpage>-513</lpage>
  <url>http://www.ncbi.nlm.nih.gov/pmc/articles/PMC2995668/</url>
</bibl>

<bibl id="B27">
  <title><p>{An overview of MetaMap: historical perspective and recent
  advances.}</p></title>
  <aug>
    <au><snm>Aronson</snm><fnm>AR</fnm></au>
    <au><snm>Lang</snm><fnm>FM</fnm></au>
  </aug>
  <source>Journal of the American Medical Informatics Association :
  JAMIA</source>
  <pubdate>2010</pubdate>
  <volume>17</volume>
  <issue>3</issue>
  <fpage>229</fpage>
  <lpage>-236</lpage>
</bibl>

<bibl id="B28">
  <title><p>Self-Supervised Detection of Contextual Synonyms in a Multi-Class
  Setting: Phenotype Annotation Use Case</p></title>
  <aug>
    <au><snm>Zhang</snm><fnm>J</fnm></au>
    <au><snm>Bolanos Trujillo</snm><fnm>L</fnm></au>
    <au><snm>Li</snm><fnm>T</fnm></au>
    <au><snm>Tanwar</snm><fnm>A</fnm></au>
    <au><snm>Freire</snm><fnm>G</fnm></au>
    <au><snm>Yang</snm><fnm>X</fnm></au>
    <au><snm>Ive</snm><fnm>J</fnm></au>
    <au><snm>Gupta</snm><fnm>V</fnm></au>
    <au><snm>Guo</snm><fnm>Y</fnm></au>
  </aug>
  <source>Proceedings of the 2021 Conference on Empirical Methods in Natural
  Language Processing</source>
  <publisher>Online and Punta Cana, Dominican Republic: Association for
  Computational Linguistics</publisher>
  <pubdate>2021</pubdate>
  <fpage>8754</fpage>
  <lpage>-8769</lpage>
  <url>https://aclanthology.org/2021.emnlp-main.690</url>
</bibl>

<bibl id="B29">
  <title><p>MIMIC-III, a freely accessible critical care database</p></title>
  <aug>
    <au><snm>Johnson</snm><fnm>AE</fnm></au>
    <au><snm>Pollard</snm><fnm>TJ</fnm></au>
    <au><snm>Shen</snm><fnm>L</fnm></au>
    <au><snm>Li Wei</snm><fnm>HL</fnm></au>
    <au><snm>Feng</snm><fnm>M</fnm></au>
    <au><snm>Ghassemi</snm><fnm>M</fnm></au>
    <au><snm>Moody</snm><fnm>B</fnm></au>
    <au><snm>Szolovits</snm><fnm>P</fnm></au>
    <au><snm>Celi</snm><fnm>LA</fnm></au>
    <au><snm>Mark</snm><fnm>RG</fnm></au>
  </aug>
  <source>Scientific data</source>
  <publisher>Nature Publishing Group</publisher>
  <pubdate>2016</pubdate>
  <volume>3</volume>
  <issue>1</issue>
  <fpage>1</fpage>
  <lpage>-9</lpage>
</bibl>

<bibl id="B30">
  <title><p>Multitask learning and benchmarking with clinical time series
  data</p></title>
  <aug>
    <au><snm>Harutyunyan</snm><fnm>H</fnm></au>
    <au><snm>Khachatrian</snm><fnm>H</fnm></au>
    <au><snm>Kale</snm><fnm>DC</fnm></au>
    <au><snm>Ver Steeg</snm><fnm>G</fnm></au>
    <au><snm>Galstyan</snm><fnm>A</fnm></au>
  </aug>
  <source>Scientific data</source>
  <publisher>Nature Publishing Group</publisher>
  <pubdate>2019</pubdate>
  <volume>6</volume>
  <issue>1</issue>
  <fpage>1</fpage>
  <lpage>-18</lpage>
</bibl>

<bibl id="B31">
  <title><p>Gene prioritization through genomic data fusion</p></title>
  <aug>
    <au><snm>Aerts</snm><fnm>S</fnm></au>
    <au><snm>Lambrechts</snm><fnm>D</fnm></au>
    <au><snm>Maity</snm><fnm>S</fnm></au>
    <au><snm>Van Loo</snm><fnm>P</fnm></au>
    <au><snm>Coessens</snm><fnm>B</fnm></au>
    <au><snm>De Smet</snm><fnm>F</fnm></au>
    <au><snm>Tranchevent</snm><fnm>LC</fnm></au>
    <au><snm>De Moor</snm><fnm>B</fnm></au>
    <au><snm>Marynen</snm><fnm>P</fnm></au>
    <au><snm>Hassan</snm><fnm>B</fnm></au>
    <au><cnm>others</cnm></au>
  </aug>
  <source>Nature biotechnology</source>
  <publisher>Nature Publishing Group</publisher>
  <pubdate>2006</pubdate>
  <volume>24</volume>
  <issue>5</issue>
  <fpage>537</fpage>
</bibl>

<bibl id="B32">
  <title><p>Deep phenotyping on electronic health records facilitates genetic
  diagnosis by clinical exomes</p></title>
  <aug>
    <au><snm>Son</snm><fnm>JH</fnm></au>
    <au><snm>Xie</snm><fnm>G</fnm></au>
    <au><snm>Yuan</snm><fnm>C</fnm></au>
    <au><snm>Ena</snm><fnm>L</fnm></au>
    <au><snm>Li</snm><fnm>Z</fnm></au>
    <au><snm>Goldstein</snm><fnm>A</fnm></au>
    <au><snm>Huang</snm><fnm>L</fnm></au>
    <au><snm>Wang</snm><fnm>L</fnm></au>
    <au><snm>Shen</snm><fnm>F</fnm></au>
    <au><snm>Liu</snm><fnm>H</fnm></au>
    <au><cnm>others</cnm></au>
  </aug>
  <source>The American Journal of Human Genetics</source>
  <publisher>Elsevier</publisher>
  <pubdate>2018</pubdate>
  <volume>103</volume>
  <issue>1</issue>
  <fpage>58</fpage>
  <lpage>-73</lpage>
</bibl>

<bibl id="B33">
  <title><p>{Ensembles of natural language processing systems for portable
  phenotyping solutions}</p></title>
  <aug>
    <au><snm>Liu</snm><fnm>C</fnm></au>
    <au><snm>Ta</snm><fnm>CN</fnm></au>
    <au><snm>Rogers</snm><fnm>JR</fnm></au>
    <au><snm>Li</snm><fnm>Z</fnm></au>
    <au><snm>Lee</snm><fnm>J</fnm></au>
    <au><snm>Butler</snm><fnm>AM</fnm></au>
    <au><snm>Shang</snm><fnm>N</fnm></au>
    <au><snm>Kury</snm><fnm>FSP</fnm></au>
    <au><snm>Wang</snm><fnm>L</fnm></au>
    <au><snm>Shen</snm><fnm>F</fnm></au>
    <au><snm>Liu</snm><fnm>H</fnm></au>
    <au><snm>Ena</snm><fnm>L</fnm></au>
    <au><snm>Friedman</snm><fnm>C</fnm></au>
    <au><snm>Weng</snm><fnm>C</fnm></au>
  </aug>
  <source>Journal of Biomedical Informatics</source>
  <pubdate>2019</pubdate>
  <volume>100</volume>
  <fpage>103318</fpage>
  <url>http://www.sciencedirect.com/science/article/pii/S1532046419302370</url>
</bibl>

<bibl id="B34">
  <title><p>{Identifying sub-phenotypes of acute kidney injury using structured
  and unstructured electronic health record data with memory
  networks.}</p></title>
  <aug>
    <au><snm>Xu</snm><fnm>Z</fnm></au>
    <au><snm>Chou</snm><fnm>J</fnm></au>
    <au><snm>Zhang</snm><fnm>XS</fnm></au>
    <au><snm>Luo</snm><fnm>Y</fnm></au>
    <au><snm>Isakova</snm><fnm>T</fnm></au>
    <au><snm>Adekkanattu</snm><fnm>P</fnm></au>
    <au><snm>Ancker</snm><fnm>JS</fnm></au>
    <au><snm>Jiang</snm><fnm>G</fnm></au>
    <au><snm>Kiefer</snm><fnm>RC</fnm></au>
    <au><snm>Pacheco</snm><fnm>JA</fnm></au>
    <au><snm>Rasmussen</snm><fnm>LV</fnm></au>
    <au><snm>Pathak</snm><fnm>J</fnm></au>
    <au><snm>Wang</snm><fnm>F</fnm></au>
  </aug>
  <source>Journal of biomedical informatics</source>
  <pubdate>2020</pubdate>
  <volume>102</volume>
  <fpage>103361</fpage>
</bibl>

<bibl id="B35">
  <title><p>{Clinical Utility of the Automatic Phenotype Annotation in
  Unstructured Clinical Notes: ICU Use Cases}</p></title>
  <aug>
    <au><snm>Zhang</snm><fnm>J</fnm></au>
    <au><snm>Bolanos</snm><fnm>L</fnm></au>
    <au><snm>Tanwar</snm><fnm>A</fnm></au>
    <au><snm>Sokol</snm><fnm>A</fnm></au>
    <au><snm>Ive</snm><fnm>J</fnm></au>
    <au><snm>Gupta</snm><fnm>V</fnm></au>
    <au><snm>Guo</snm><fnm>Y</fnm></au>
  </aug>
  <source>arXiv preprint arXiv:2107.11665</source>
  <pubdate>2021</pubdate>
</bibl>

<bibl id="B36">
  <title><p>{The Human Phenotype Ontology in 2021}</p></title>
  <aug>
    <au><snm>K{\"{o}}hler</snm><fnm>S</fnm></au>
    <au><snm>Gargano</snm><fnm>MA</fnm></au>
    <au><snm>Matentzoglu</snm><fnm>N</fnm></au>
    <au><snm>Carmody</snm><fnm>L</fnm></au>
    <au><snm>Lewis{-}Smith</snm><fnm>D</fnm></au>
    <au><snm>Vasilevsky</snm><fnm>NA</fnm></au>
    <au><snm>Danis</snm><fnm>D</fnm></au>
    <au><snm>Balagura</snm><fnm>G</fnm></au>
    <au><snm>Baynam</snm><fnm>G</fnm></au>
    <au><snm>Brower</snm><fnm>AM</fnm></au>
    <au><snm>Callahan</snm><fnm>TJ</fnm></au>
    <au><snm>Chute</snm><fnm>CG</fnm></au>
    <au><snm>Est</snm><fnm>JL</fnm></au>
    <au><snm>Galer</snm><fnm>PD</fnm></au>
    <au><snm>Ganesan</snm><fnm>S</fnm></au>
    <au><snm>Griese</snm><fnm>M</fnm></au>
    <au><snm>Haimel</snm><fnm>M</fnm></au>
    <au><snm>Pazmandi</snm><fnm>J</fnm></au>
    <au><snm>Hanauer</snm><fnm>M</fnm></au>
    <au><snm>Harris</snm><fnm>NL</fnm></au>
    <au><snm>Hartnett</snm><fnm>M</fnm></au>
    <au><snm>Hastreiter</snm><fnm>M</fnm></au>
    <au><snm>Hauck</snm><fnm>F</fnm></au>
    <au><snm>He</snm><fnm>Y</fnm></au>
    <au><snm>Jeske</snm><fnm>T</fnm></au>
    <au><snm>Kearney</snm><fnm>H</fnm></au>
    <au><snm>Kindle</snm><fnm>G</fnm></au>
    <au><snm>Klein</snm><fnm>C</fnm></au>
    <au><snm>Knoflach</snm><fnm>K</fnm></au>
    <au><snm>Krause</snm><fnm>R</fnm></au>
    <au><snm>Lagorce</snm><fnm>D</fnm></au>
    <au><snm>McMurry</snm><fnm>JA</fnm></au>
    <au><snm>Miller</snm><fnm>JA</fnm></au>
    <au><snm>Munoz{-}Torres</snm><fnm>MC</fnm></au>
    <au><snm>Peters</snm><fnm>RL</fnm></au>
    <au><snm>Rapp</snm><fnm>CK</fnm></au>
    <au><snm>Rath</snm><fnm>A</fnm></au>
    <au><snm>Rind</snm><fnm>SA</fnm></au>
    <au><snm>Rosenberg</snm><fnm>AZ</fnm></au>
    <au><snm>Segal</snm><fnm>MM</fnm></au>
    <au><snm>Seidel</snm><fnm>MG</fnm></au>
    <au><snm>Smedley</snm><fnm>D</fnm></au>
    <au><snm>Talmy</snm><fnm>T</fnm></au>
    <au><snm>Thomas</snm><fnm>Y</fnm></au>
    <au><snm>Wiafe</snm><fnm>SA</fnm></au>
    <au><snm>Xian</snm><fnm>J</fnm></au>
    <au><snm>Y{\"{u}}ksel</snm><fnm>Z</fnm></au>
    <au><snm>Helbig</snm><fnm>I</fnm></au>
    <au><snm>Mungall</snm><fnm>CJ</fnm></au>
    <au><snm>Haendel</snm><fnm>MA</fnm></au>
    <au><snm>Robinson</snm><fnm>PN</fnm></au>
  </aug>
  <source>Nucleic Acids Res.</source>
  <pubdate>2021</pubdate>
  <volume>49</volume>
  <issue>Database-Issue</issue>
  <fpage>D1207</fpage>
  <lpage>-D1217</lpage>
  <url>https://doi.org/10.1093/nar/gkaa1043</url>
</bibl>

<bibl id="B37">
  <title><p>Attention is All you Need</p></title>
  <aug>
    <au><snm>Vaswani</snm><fnm>A</fnm></au>
    <au><snm>Shazeer</snm><fnm>N</fnm></au>
    <au><snm>Parmar</snm><fnm>N</fnm></au>
    <au><snm>Uszkoreit</snm><fnm>J</fnm></au>
    <au><snm>Jones</snm><fnm>L</fnm></au>
    <au><snm>Gomez</snm><fnm>AN</fnm></au>
    <au><snm>Kaiser</snm><fnm>\L{ukasz}</fnm></au>
    <au><snm>Polosukhin</snm><fnm>I</fnm></au>
  </aug>
  <source>Advances in Neural Information Processing Systems 30</source>
  <pubdate>2017</pubdate>
  <fpage>5998</fpage>
  <lpage>-6008</lpage>
  <url>http://papers.nips.cc/paper/7181-attention-is-all-you-need.pdf</url>
</bibl>

<bibl id="B38">
  <title><p>MetaMap Lite: an evaluation of a new Java implementation of
  MetaMap</p></title>
  <aug>
    <au><snm>Demner Fushman</snm><fnm>D</fnm></au>
    <au><snm>Rogers</snm><fnm>WJ</fnm></au>
    <au><snm>Aronson</snm><fnm>AR</fnm></au>
  </aug>
  <source>Journal of the American Medical Informatics Association</source>
  <publisher>Oxford University Press</publisher>
  <pubdate>2017</pubdate>
  <volume>24</volume>
  <issue>4</issue>
  <fpage>841</fpage>
  <lpage>-844</lpage>
</bibl>

<bibl id="B39">
  <title><p>{Identifying clinical terms in medical text using Ontology-Guided
  machine learning}</p></title>
  <aug>
    <au><snm>Arbabi</snm><fnm>A</fnm></au>
    <au><snm>Adams</snm><fnm>DR</fnm></au>
    <au><snm>Fidler</snm><fnm>S</fnm></au>
    <au><snm>Brudno</snm><fnm>M</fnm></au>
  </aug>
  <source>JMIR medical informatics</source>
  <publisher>JMIR Publications Inc., Toronto, Canada</publisher>
  <pubdate>2019</pubdate>
  <volume>7</volume>
  <issue>2</issue>
  <fpage>e12596</fpage>
</bibl>

<bibl id="B40">
  <title><p>MedCAT -- Medical Concept Annotation Tool</p></title>
  <aug>
    <au><snm>Kraljevic</snm><fnm>Z</fnm></au>
    <au><snm>Bean</snm><fnm>D</fnm></au>
    <au><snm>Mascio</snm><fnm>A</fnm></au>
    <au><snm>Roguski</snm><fnm>L</fnm></au>
    <au><snm>Folarin</snm><fnm>A</fnm></au>
    <au><snm>Roberts</snm><fnm>A</fnm></au>
    <au><snm>Bendayan</snm><fnm>R</fnm></au>
    <au><snm>Dobson</snm><fnm>R</fnm></au>
  </aug>
  <pubdate>2019</pubdate>
</bibl>

<bibl id="B41">
  <title><p>{BERT}: Pre-training of Deep Bidirectional Transformers for
  Language Understanding</p></title>
  <aug>
    <au><snm>Devlin</snm><fnm>J</fnm></au>
    <au><snm>Chang</snm><fnm>MW</fnm></au>
    <au><snm>Lee</snm><fnm>K</fnm></au>
    <au><snm>Toutanova</snm><fnm>K</fnm></au>
  </aug>
  <source>Proceedings of the 2019 Conference of the North {A}merican Chapter of
  the Association for Computational Linguistics: Human Language Technologies,
  Volume 1 (Long and Short Papers)</source>
  <publisher>Minneapolis, Minnesota: Association for Computational
  Linguistics</publisher>
  <pubdate>2019</pubdate>
  <fpage>4171</fpage>
  <lpage>-4186</lpage>
  <url>https://aclanthology.org/N19-1423</url>
</bibl>

<bibl id="B42">
  <title><p>{BioBERT: a pre-trained biomedical language representation model
  for biomedical text mining}</p></title>
  <aug>
    <au><snm>Lee</snm><fnm>J</fnm></au>
    <au><snm>Yoon</snm><fnm>W</fnm></au>
    <au><snm>Kim</snm><fnm>S</fnm></au>
    <au><snm>Kim</snm><fnm>D</fnm></au>
    <au><snm>Kim</snm><fnm>S</fnm></au>
    <au><snm>So</snm><fnm>CH</fnm></au>
    <au><snm>Kang</snm><fnm>J</fnm></au>
  </aug>
  <source>Bioinformatics</source>
  <pubdate>2019</pubdate>
  <url>https://doi.org/10.1093/bioinformatics/btz682</url>
</bibl>

<bibl id="B43">
  <title><p>Publicly Available Clinical {BERT} Embeddings</p></title>
  <aug>
    <au><snm>Alsentzer</snm><fnm>E</fnm></au>
    <au><snm>Murphy</snm><fnm>J</fnm></au>
    <au><snm>Boag</snm><fnm>W</fnm></au>
    <au><snm>Weng</snm><fnm>WH</fnm></au>
    <au><snm>Jindi</snm><fnm>D</fnm></au>
    <au><snm>Naumann</snm><fnm>T</fnm></au>
    <au><snm>McDermott</snm><fnm>M</fnm></au>
  </aug>
  <source>Proceedings of the 2nd Clinical Natural Language Processing
  Workshop</source>
  <publisher>Minneapolis, Minnesota, USA: Association for Computational
  Linguistics</publisher>
  <pubdate>2019</pubdate>
  <fpage>72</fpage>
  <lpage>-78</lpage>
  <url>https://aclanthology.org/W19-1909</url>
</bibl>

<bibl id="B44">
  <title><p>{S}ci{BERT}: A Pretrained Language Model for Scientific
  Text</p></title>
  <aug>
    <au><snm>Beltagy</snm><fnm>I</fnm></au>
    <au><snm>Lo</snm><fnm>K</fnm></au>
    <au><snm>Cohan</snm><fnm>A</fnm></au>
  </aug>
  <source>Proceedings of the 2019 Conference on Empirical Methods in Natural
  Language Processing and the 9th International Joint Conference on Natural
  Language Processing (EMNLP-IJCNLP)</source>
  <publisher>Hong Kong, China: Association for Computational
  Linguistics</publisher>
  <pubdate>2019</pubdate>
  <fpage>3615</fpage>
  <lpage>-3620</lpage>
  <url>https://aclanthology.org/D19-1371</url>
</bibl>

<bibl id="B45">
  <title><p>AutoML: A Survey of the State-of-the-Art</p></title>
  <aug>
    <au><snm>He</snm><fnm>X</fnm></au>
    <au><snm>Zhao</snm><fnm>K</fnm></au>
    <au><snm>Chu</snm><fnm>X</fnm></au>
  </aug>
  <source>Knowledge-Based Systems</source>
  <publisher>Elsevier</publisher>
  <pubdate>2021</pubdate>
  <volume>212</volume>
  <fpage>106622</fpage>
</bibl>

<bibl id="B46">
  <title><p>{BOHB}: Robust and Efficient Hyperparameter Optimization at
  Scale</p></title>
  <aug>
    <au><snm>Falkner</snm><fnm>S</fnm></au>
    <au><snm>Klein</snm><fnm>A</fnm></au>
    <au><snm>Hutter</snm><fnm>F</fnm></au>
  </aug>
  <source>Proceedings of the 35th International Conference on Machine
  Learning</source>
  <pubdate>2018</pubdate>
  <fpage>1436</fpage>
  <lpage>-1445</lpage>
</bibl>

<bibl id="B47">
  <title><p>Scikit-learn: Machine Learning in {P}ython</p></title>
  <aug>
    <au><snm>Pedregosa</snm><fnm>F.</fnm></au>
    <au><snm>Varoquaux</snm><fnm>G.</fnm></au>
    <au><snm>Gramfort</snm><fnm>A.</fnm></au>
    <au><snm>Michel</snm><fnm>V.</fnm></au>
    <au><snm>Thirion</snm><fnm>B.</fnm></au>
    <au><snm>Grisel</snm><fnm>O.</fnm></au>
    <au><snm>Blondel</snm><fnm>M.</fnm></au>
    <au><snm>Prettenhofer</snm><fnm>P.</fnm></au>
    <au><snm>Weiss</snm><fnm>R.</fnm></au>
    <au><snm>Dubourg</snm><fnm>V.</fnm></au>
    <au><snm>Vanderplas</snm><fnm>J.</fnm></au>
    <au><snm>Passos</snm><fnm>A.</fnm></au>
    <au><snm>Cournapeau</snm><fnm>D.</fnm></au>
    <au><snm>Brucher</snm><fnm>M.</fnm></au>
    <au><snm>Perrot</snm><fnm>M.</fnm></au>
    <au><snm>Duchesnay</snm><fnm>E.</fnm></au>
  </aug>
  <source>Journal of Machine Learning Research</source>
  <pubdate>2011</pubdate>
  <volume>12</volume>
  <fpage>2825</fpage>
  <lpage>-2830</lpage>
</bibl>

<bibl id="B48">
  <title><p>A unified approach to interpreting model predictions</p></title>
  <aug>
    <au><snm>Lundberg</snm><fnm>SM</fnm></au>
    <au><snm>Lee</snm><fnm>SI</fnm></au>
  </aug>
  <source>Advances in neural information processing systems</source>
  <pubdate>2017</pubdate>
  <volume>30</volume>
</bibl>

<bibl id="B49">
  <title><p>Random forests</p></title>
  <aug>
    <au><snm>Breiman</snm><fnm>L</fnm></au>
  </aug>
  <source>Machine Learning</source>
  <publisher>Springer</publisher>
  <pubdate>2001</pubdate>
  <volume>45</volume>
  <fpage>5</fpage>
  <lpage>32</lpage>
  <url>https://link.springer.com/article/10.1023/A:1010933404324</url>
</bibl>

</refgrp>
} 

\appendix

\end{document}